\documentclass{article}

\usepackage[preprint]{corl_2026} % Uncomment for pre-prints (e.g., arxiv); This is like ``final'', but will remove the CORL footnote.

\usepackage[numbers]{natbib}
\usepackage{multicol}
\usepackage{hyperref}
\usepackage{comment}
\usepackage{siunitx}
\usepackage{relsize}
\usepackage{ifthen}
\usepackage[colorinlistoftodos]{todonotes}
\usepackage[vlined,ruled,linesnumbered]{algorithm2e}
\usepackage{graphics} % for pdf, bitmapped graphics files
\usepackage{rotating}
\usepackage{color}
\usepackage{enumerate}
\usepackage[T1]{fontenc}
\usepackage{psfrag}
\usepackage{epsfig} % for postscript graphics files
\usepackage{booktabs}
\usepackage{graphicx}
\usepackage{url}
\usepackage{multirow}
\usepackage{array}
\usepackage{latexsym}
\usepackage{amsfonts}
\usepackage{amsmath}
\usepackage{amssymb}
\usepackage{dsfont}
\usepackage{xstring}
\usepackage{algorithmic}
\usepackage{multirow}
\usepackage{xcolor}
\usepackage{prettyref}
\usepackage{flexisym}
\usepackage{bigdelim}
\usepackage{breqn} % load this l\usepackage{multicol}
\usepackage{comment}
\usepackage{listings}

\usepackage{xspace}
\usepackage{bm}
\usepackage{soul}
\graphicspath{{./figures/}}
\usepackage{tikz}
\usetikzlibrary{matrix,calc}
\usepackage{lipsum}
\usepackage{mdwlist}

\makecompactlist{itemize}{stditemize}
\usepackage{enumitem}
\usepackage{caption}
\usepackage{epstopdf}
\usepackage{subfigure}

\usepackage{amsthm}
\usepackage{mathtools}
\usepackage{cleveref}
\usepackage{appendix}

\usepackage{import}
\usepackage{svg}

\PassOptionsToPackage{end}{algorithmic}

\newtheorem{problem}{Problem}

\newtheorem{proposition}{Proposition}

\newtheorem{remark}{Remark}
\newtheorem{example}{Example}

\newcommand{\bdmath}{\begin{dmath}}
\newcommand{\edmath}{\end{dmath}}
\newcommand{\beq}{\begin{equation}}
\newcommand{\eeq}{\end{equation}}
\newcommand{\bdm}{\begin{displaymath}}
\newcommand{\edm}{\end{displaymath}}
\newcommand{\bea}{\begin{eqnarray}}
\newcommand{\eea}{\end{eqnarray}}
\newcommand{\beal}{\beq \begin{array}{lll}}
\newcommand{\eeal}{\end{array} \eeq}
\newcommand{\beas}{\begin{eqnarray*}}
\newcommand{\eeas}{\end{eqnarray*}}
\newcommand{\ba}{\begin{array}}
\newcommand{\ea}{\end{array}}
\newcommand{\bit}{\begin{itemize}}
\newcommand{\eit}{\end{itemize}}
\newcommand{\ben}{\begin{enumerate}}
\newcommand{\een}{\end{enumerate}}

\newcommand{\calD}{{\cal D}}

\newcommand{\calF}{{\cal F}}

\newcommand{\calH}{{\cal H}}

\newcommand{\calO}{{\cal O}}

\definecolor{myblue}{RGB}{65 105 225}

\newcommand{\hide}[1]{}

\newcommand{\hiddenText}{{\color{gray} hidden text.}}
\newcommand{\hideWithText}[1]{\hiddenText}

\newcommand{\ie}{\emph{i.e.},\xspace}
\newcommand{\eg}{\emph{e.g.},\xspace}

\newcommand{\blue}[1]{{\color{blue}#1}}
\title{Efficient Domain-Adaptive Policy Learning via Kernel Representation with Application to Quadrotor Control under Non-Stationary Disturbances}
\author{
  Hongyu Zhou\\
  University of Michigan, Ann Arbor \\
  \texttt{zhouhy@umich.edu} \\
  \And
  Mingtian Tan\\
  University of Michigan, Ann Arbor \\
  \texttt{teemty@umich.edu} \\
  \And
  Vasileios Tzoumas\\
  University of Michigan, Ann Arbor \\
  \texttt{vtzoumas@umich.edu} \\}

\begin{document}
	\maketitle
	
	\begin{abstract}
We present an algorithm for efficient domain-adaptive policy learning via kernel representations. 
Learning domain-adaptive policies is challenging since it requires an environment representation that is both sufficiently expressive to model complex sim-to-real gaps during offline training, and computationally efficient enough to support rapid online adaptation during deployment. 
For instance, a quadrotor may encounter time-varying, non-stationary disturbances, such as sudden gusts of wind, payload shifts, or transitions between distinct flight regimes with and without ground effects.  
To address these challenges, we model unknown disturbances using a differentiable kernel approximation based on random Fourier features. During the offline training phase, we randomly sample kernel coefficients and bandwidth parameters to generate a rich diversity of disturbance profiles. We then optimize the control policy via differentiable simulation with analytical gradients, a process that takes only $50$ seconds of training time on an RTX 4090 GPU. 
During hardware deployment, the policy adapts to non-stationary environments in real time by updating both the kernel coefficients and bandwidth through online least-squares estimation.
We evaluate our method on quadrotor trajectory tracking tasks across high-fidelity numerical simulations and hardware experiments using Crazyflie, subjected to various disturbances, including complex aerodynamic effects, wind, ground effects, and payload fluctuations. 
% The experimental results demonstrate that our approach consistently outperforms state-of-the-art learning-based and model-based adaptive control baselines. 
\end{abstract}


	\section{Introduction}
\label{sec:intro}
In the future, mobile robots will perform agile tasks such as aerial transportation, search and rescue, and inspection in cluttered environments. Such tasks require accurate and efficient tracking control of the robot's motion. But achieving accuracy and efficiency is challenging since such tasks often require the robot to operate under uncertain conditions, particularly, under unknown disturbances that may be time-varying and non-stationary. For example, a quadrotor may encounter sudden gusts of wind, payload shifts that change the center of mass mid-flight, or transitions between distinct flight regimes such as moving in and out of ground effect during takeoff and landing. We are motivated by the future of autonomy where robots will autonomously perform such tasks despite these unknown, non-stationary disturbances. Achieving this requires an environment representation that is sufficiently expressive to model complex sim-to-real gaps during offline policy training, and computationally efficient enough to support rapid online adaptation during deployment.

Existing methods do not satisfy both requirements simultaneously. 
Robust control~\cite{zhou1998essentials,mayne2005robust,zhou2023safe,martin2024guarantees,liu2024robust} selects control inputs assuming a worst-case realization of disturbances, which can be conservative. Classical adaptive control~\cite{slotine1991applied,krstic1995nonlinear,ioannou1996robust} often assumes parametric uncertainty additive to the known system dynamics, where unknown coefficients multiply a fixed set of known basis functions. The coefficients are updated online to compensate for the estimated disturbances. However, the capability of disturbance compensation is limited by the expressiveness of the chosen basis functions. Learning-based control~\cite{kumar2021rma,xue2026robust,zhang2025learning,huang2023datt} trains a neural-network policy conditioned on the output of an environment encoder to achieve domain adaptation. However, the encoder is fixed after offline training and cannot represent disturbances beyond the diversity of the training environments. Online learning methods~\cite{hazan2022introduction,agarwal2019online,zhao2022non,zhou2023safecdc} select control inputs based on past information alone and are agnostic to the form of the disturbance, but in practice, they are sensitive to the choice of tuning parameters~\cite{zhou2023safecdc}. 
% The closest prior work~\cite{zhou2025simultaneous} uses random Fourier features (RFF) for model predictive control and updates the feature coefficients online, but the kernel bandwidth is fixed at training time. As a result, the achievable approximation is restricted by the expressiveness of a single, pre-committed kernel bandwidth. 
We further discuss related work in Appendix~\ref{sec:related}.

To address these challenges, we provide an algorithm for efficient domain-adaptive policy learning via kernel representations~(\Cref{fig:framework}). We model the unknown disturbance with a differentiable kernel approximation based on random Fourier features~(RFF)~\cite{rahimi2007random,rahimi2008uniform}. The approximation has two sets of parameters: the kernel coefficients and the bandwidth. Both parameters are learnable and the function approximation is differentiable, allowing efficient offline policy learning and online adaptation. Specifically, the kernel approximation based on RFF enables approximating functions in Reproducing Kernel Hilbert Space, providing a rich representation for modeling complex sim-to-real gaps. In addition, the basis functions can be adapted by adjusting the kernel bandwidth parameters, overcoming the limitations of a fixed set of basis functions.

The proposed algorithm is composed of offline learning and online adaptation phases. In the offline phase, we randomly sample the kernel coefficients and the bandwidth to generate a rich diversity of disturbance profiles, and we optimize the policy via differentiable simulation with analytical gradients. The offline training takes only $50$ seconds on an RTX 4090 GPU. In the online phase, the policy is deployed zero-shot, and the kernel coefficients and the bandwidth are updated jointly at every control step through online least-squares estimation. Updating both the kernel coefficients and the bandwidth allows the basis itself to adapt to the disturbance encountered at deployment, rather than only the coefficients projected onto a pre-committed basis.

We validate our algorithm in both numerical simulations and hardware experiments. The simulations use a high-fidelity aerodynamic model together with additional disturbance types, including sinusoidal, switching (from sinusoidal to constant), and quadratic-phase sinusoidal. The hardware experiments use Crazyflie drones under various disturbances, including an unknown suspended payload that weighs $20\%$ of the drone mass and may swing, on-and-off-switching ground effects, and time-varying wind. Across all settings, our method consistently outperforms state-of-the-art learning-based and model-based adaptive control baselines~\cite{huang2023datt,zhou2025simultaneous}, improving position tracking by up to $60 - 70\%$ in simulation and up to $30\%$ on hardware.

\begin{figure*}[t]
    \centering
    \subfigure[Unknown Payload.]{\includegraphics[width=0.3\textwidth]{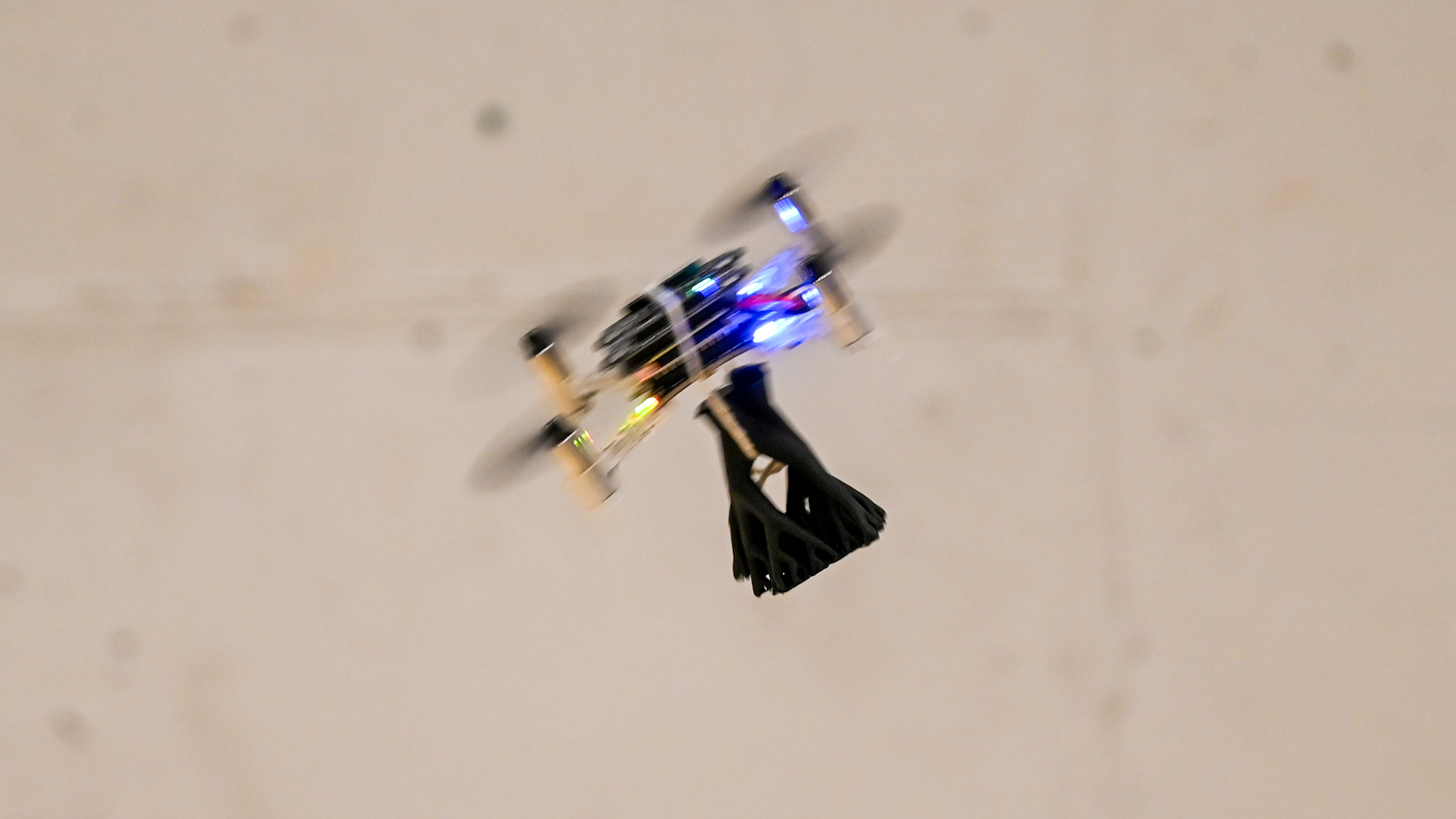}\label{fig:hw-payload}}	
    \subfigure[Ground Effect.]{\includegraphics[width=0.3\textwidth]{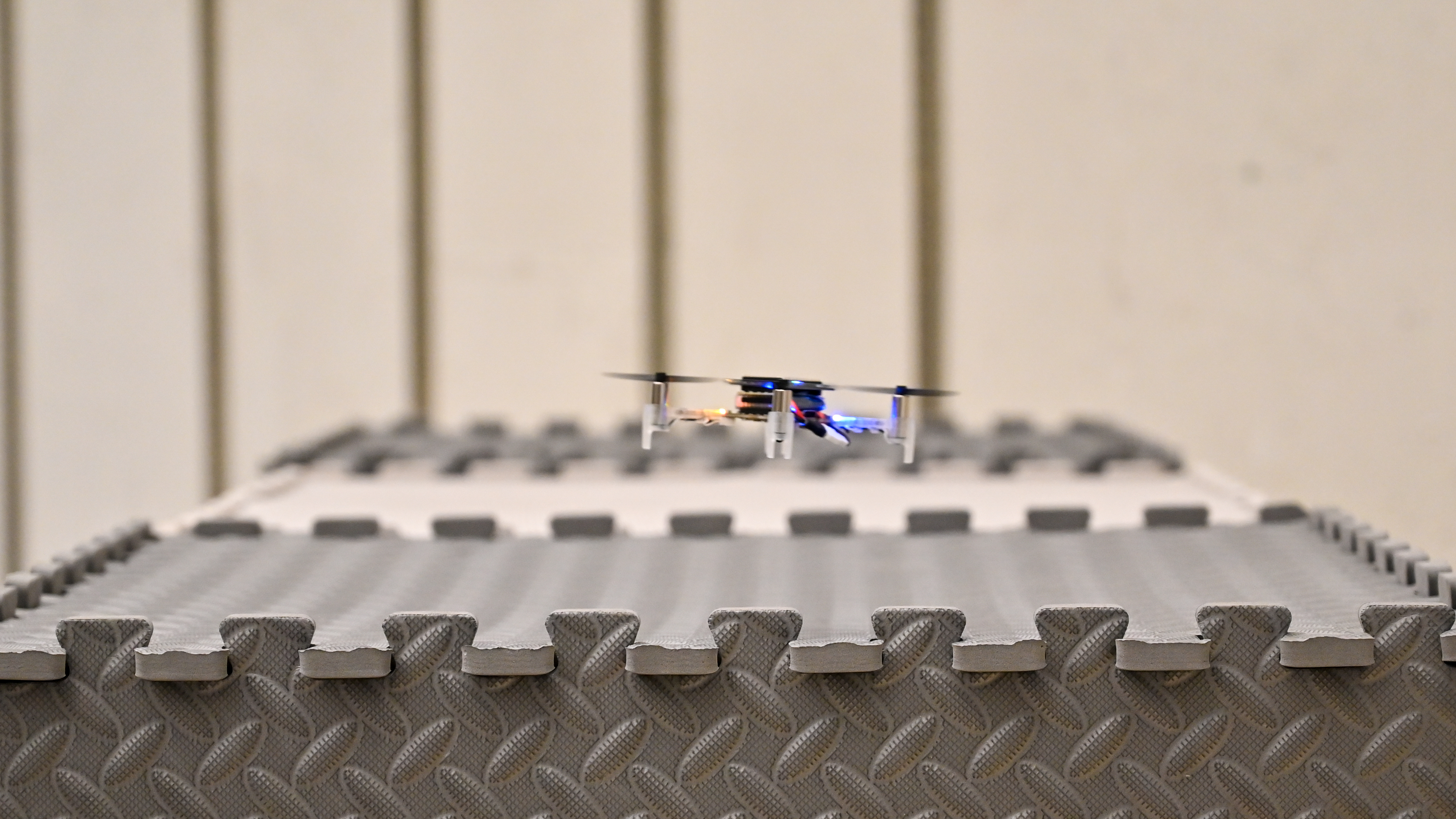}\label{fig:hw-ground}}	
    \subfigure[Wind Disturbance.]{\includegraphics[width=0.3\textwidth]{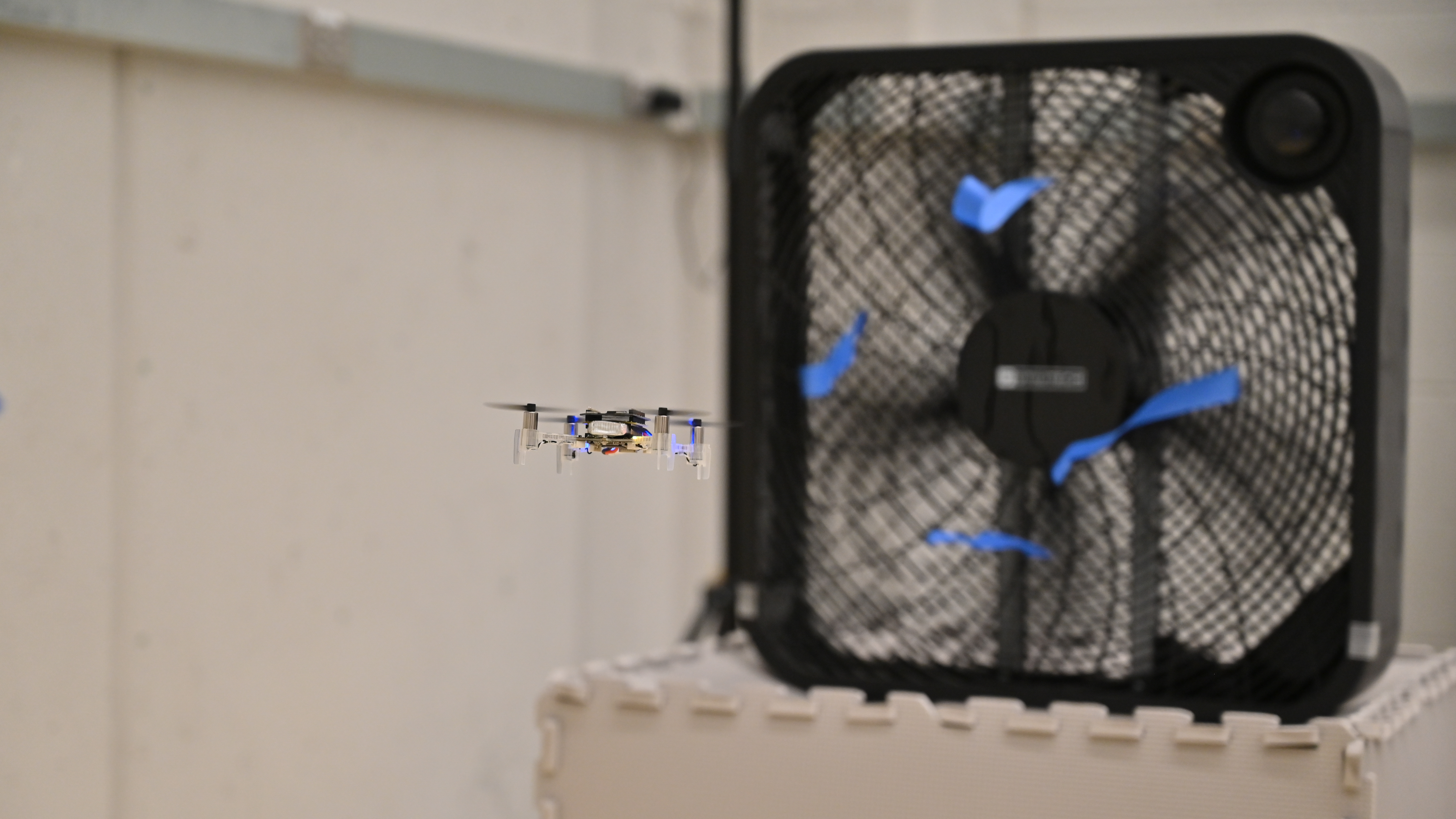}\label{fig:hw-wind}}	
    \vspace{-3mm}
    \caption{\textbf{Efficient Domain-Adaptive Policy Learning via Kernel Representation for Quadrotor Control under Non-Stationary Disturbances.} We leverage kernel representations for efficient domain-adaptive policy learning that achieves zero-shot sim-2-real transfer for quadrotor control under non-stationary disturbances: (a) unknown suspended payload that weighs $20\%$ of drone mass and may swing, (b) on-and-off-switching ground effects, and (c) time-varying wind disturbances.}
    \label{fig:hw}
    \vspace{-4mm}
\end{figure*}
	\section{Problem Formulation}
\label{sec:problem}

Consider the quadrotor dynamics as follows,
\vspace{-3mm}
\begin{equation}
    \dot{\boldsymbol{p}} = \boldsymbol{v}, \;\;  m \dot{\boldsymbol{v}} = m \boldsymbol{g} + \boldsymbol{f} + \boldsymbol{h}, \;\; \dot{\boldsymbol{q}} =\frac{1}{2}\boldsymbol{q}\otimes\left[
        0, \;
        \boldsymbol{\omega}\right]^\top,
    \label{eq:quadrotor}
\end{equation}
where $\boldsymbol{p} \in \mathbb{R}^{3}$ and $\boldsymbol{v} \in \mathbb{R}^{3}$ are position and velocity in the inertial frame, $\boldsymbol{q}$ is the quaternion, $\otimes$ is the quaternion multiplication operator, $\boldsymbol{\omega} \in \mathbb{R}^{3}$ is the body angular velocity, $m$ is the quadrotor mass, $\mathcal{J}$ is the inertia matrix of the quadrotor, $\boldsymbol{g}$ is the gravity vector, $\boldsymbol{f} =  \boldsymbol{R}\left[0\ 0\ T\right]^\top\in \mathbb{R}^3$ is the total thrust from the four rotors, $T$ is the thrust from the four rotors along the $z-$axis of the body frame, and $\boldsymbol{h} \in \mathbb{R}^3$ is the unknown disturbance force. The control input is $[T, \; \boldsymbol{\omega}^\top]^\top$. We adopt this choice of control input since common low-level flight controllers, such as PX4\footnote{\url{https://docs.px4.io/main/en/index.html}} and Betaflight\footnote{\url{https://betaflight.com/}}, accept $[T, \; \boldsymbol{\omega}^\top]^\top$ as a high-level control command.

The quadrotor dynamics in \cref{eq:quadrotor} can be compactly written as 
\begin{equation}
    \boldsymbol{x}_{t+1} = \boldsymbol{f}(\boldsymbol{x}_{t}) + \boldsymbol{g}(\boldsymbol{x}_{t}) \boldsymbol{u}_{t} + \boldsymbol{h},
\end{equation}
where $\boldsymbol{x} \triangleq [\boldsymbol{p}^\top, \; \boldsymbol{v}^\top, \; \boldsymbol{q}^\top]^\top$ and $\boldsymbol{u}  \triangleq [T, \; \boldsymbol{\omega}^\top]^\top$.

\begin{problem}[]\label{prob:control}
Given a reference trajectory $\boldsymbol{x}_{ref}$, the goal is to learn a control policy that adapts online using the estimated disturbance $\hat{\boldsymbol{h}}$ of the unknown disturbance $\boldsymbol{h}$, where the disturbance $\boldsymbol{h}$ may be time-varying or non-stationary.
\end{problem}
	\section{Algorithm}
\label{sec:algo}

\begin{figure}[t]
    \centering
    \includegraphics[width=\textwidth]{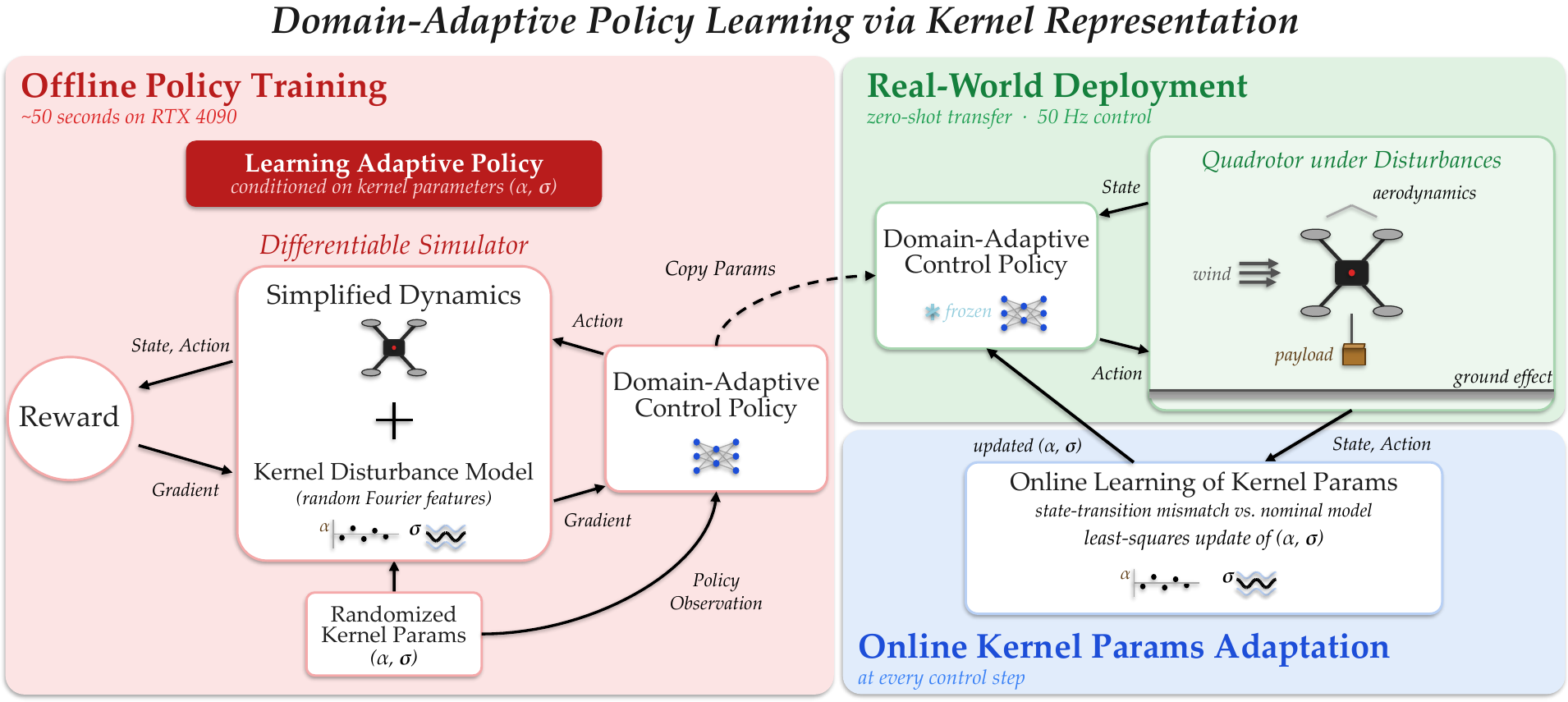}
    \caption{\textbf{Efficient Domain-Adaptive Policy Learning via Kernel Representation.} The framework is composed of offline and online phases. During the offline training phase, we randomly sample kernel coefficients and bandwidth parameters to generate a rich diversity of disturbance profiles. We then optimize the control policy via differentiable simulation with analytical gradients, a process that takes only $50$ seconds of training time on an RTX 4090 GPU. During hardware deployment, the policy adapts to non-stationary environments in real time by updating both the kernel coefficients and bandwidth parameters simultaneously through online least-squares estimation.}
    \label{fig:framework}
\end{figure}

In this section, we introduce our algorithm. We first introduce kernel representation for function approximation based on random Fourier feature~(RFF) in \Cref{subsec:RFF}, the policy optimization method with analytical gradients in \Cref{subsec:BPTT}, and online learning of kernel parameters in \Cref{subsec:OLS}. Finally, we present our algorithm in \Cref{subsec:algo} that is composed of offline and online phases~\Cref{fig:framework}: (i) offline phase utilizes the policy optimization method with analytical gradients for control policy training, and (ii) online learning of the disturbance function represented by RFF.

\subsection{Function Approximation via Random Fourier Features}\label{subsec:RFF}

We overview the randomized approximation algorithm in~\cite{boffi2022nonparametric} for approximating an $\boldsymbol{h}\left(\cdot\right)$.
The algorithm is based on RFF~\cite{rahimi2007random,rahimi2008uniform} and their extension to vector-valued functions~\cite{brault2016random,minh2016operator}.
By being randomized, the algorithm is computationally efficient
while retaining the expressiveness of the RKHS with high probability.
RFF can be viewed as linearizations of neural networks~\cite{ghorbani2021linearized,jacot2018neural}. Neural networks, in principle, can perform better than kernel methods due to greater expressivity.  However, using neural networks poses challenges in such online learning settings due to their data-hungry nature. 
Therefore, we utilize RFF to balance computational efficiency and expressiveness.

Based on the assumptions of $\boldsymbol{h}: \mathbb{R}^{d_z} \rightarrow \mathbb{R}^{d_x}$ lies in a subspace of a Reproducing Kernel Hilbert Space (RKHS) $\calH$~\cite{bach2017breaking} and the Operator-Valued Bochner's Theorem~\cite{brault2016random}, 
$\boldsymbol{h}$ can be written as $\boldsymbol{h}\left(\cdot\right) = \int_\Theta \boldsymbol{\Phi}_\nu\left(\cdot, \boldsymbol{\theta}\right) \boldsymbol{\alpha}(\boldsymbol{\theta}) \mathrm{d}\nu (\boldsymbol{\theta})$~\cite{bach2017breaking}
and there exists a finite-dimensional approximation of $\boldsymbol{h}\left(\cdot\right)$ by $\boldsymbol{h}\left(\cdot\right) \approx \hat{\boldsymbol{h}}(\cdot;\boldsymbol{\alpha})\triangleq\frac{1}{M} \sum_{i=1}^{M} \boldsymbol{\Phi}_\nu\left(\cdot, \boldsymbol{\theta}_i \right) \boldsymbol{\alpha}_i,$
where  $\boldsymbol{\Phi}_\nu \left(\boldsymbol{z}, \boldsymbol{\theta}\right)= \boldsymbol{B}(\boldsymbol{w}) \phi\left(\boldsymbol{w}^{\top} \boldsymbol{z}+b\right)$ is the feature map, 
$\boldsymbol{B}: \mathbb{R}^{d_z} \rightarrow \mathbb{R}^{d_x \times d_{1}}$, $\phi: \mathbb{R} \rightarrow[-1,1]$ is a $1$-Lipschitz function, 
$d_1 \leq d_x$, 
$\boldsymbol{\theta}_i \sim \nu$ are drawn i.i.d. from the base measure $\nu$ with $\boldsymbol{\theta}=\left(\boldsymbol{w}, b\right)$, $\boldsymbol{w} \in \mathbb{R}^{d_z}$, and $b \in \mathbb{R}$, 
$\boldsymbol{\alpha}_i \triangleq \boldsymbol{\alpha}\left(\theta_i\right)$ are parameters to be learned, and $M$ is the number of sampling points that decides the approximation accuracy.

The following shows the expressiveness of the  finite-dimensional approximation of $\boldsymbol{h}\left(\cdot\right)$, considering $\boldsymbol{\alpha}_{i} \in \calD$, where 
$\calD\triangleq \{ \boldsymbol{\alpha} \mid \|\boldsymbol{\alpha}\| \leq B_h\}$.

\begin{proposition}[Uniformly Approximation Error~\cite{boffi2022nonparametric}]\label{prop:approx_error}
     Given the base measure $\nu$ and assume $\boldsymbol{h} \in$ $\mathcal{F}_{2}\left(B_{{h}}\right)$, where $\calF_2 \left(B_h\right) \triangleq  \Bigg\{ \boldsymbol{h}\left(\cdot\right) = \left. \int_\Theta \boldsymbol{\Phi}_\nu\left(\cdot, \boldsymbol{\theta}\right) \boldsymbol{\alpha}(\boldsymbol{\theta}) \mathrm{d}\nu (\boldsymbol{\theta})  \right\vert \boldsymbol{\alpha} \in \calD \Bigg\}.$ 
     Let $\delta \in(0,1)$.  With probability at least $1-\delta$, there exist $\left\{\boldsymbol{\alpha}_{i}\right\}_{i=1}^{M} \in \calD$, \ie $\|\boldsymbol{\alpha}_{i}\| \leq B_h$, such that
\begin{equation} 
        \left\| \boldsymbol{h}\left(\cdot\right) - \frac{1}{M} \sum_{i=1}^{M} \boldsymbol{\Phi}_\nu\left(\cdot, \boldsymbol{\theta}_{i}\right) \boldsymbol{\alpha}_{i}\right\|_{\infty} \leq \calO\left(\frac{1}{\sqrt{M}}\right).
    \end{equation}
\end{proposition}
    
Therefore, \Cref{prop:approx_error} indicates that the uniformly approximation error scales $\calO\left(\frac{1}{\sqrt{M}}\right)$ given the base measure $\nu$. 

\begin{example}[Random Fourier Features with Gaussian Kernel]
    Consider Gaussian Kernel $\boldsymbol{K}(\boldsymbol{x}, \boldsymbol{y}) = \exp \left( -\frac{\|\boldsymbol{x}- \boldsymbol{y} \|^2}{2\sigma^2} \right)$ with bandwidth $\sigma$. Each entry of $\boldsymbol{\Phi}_\nu$, \ie $\boldsymbol{\Phi}_{ij}$, is $\cos\left( \sigma \boldsymbol{w}_{ij}^\top \boldsymbol{z} + b_{ij}\right)$, where the base measure $\nu$ is a Gaussian distribution $\mathcal{N}\left(\boldsymbol{0}, \sigma^2\boldsymbol{I}\right)$, $\boldsymbol{w}_{ij}$ is sampled from standard Gaussian distribution, and $b$ is sampled from uniform distribution of $[0,\; 2\pi]$.
\end{example}

\begin{remark}[Importance of Kernel Bandwidth]
    Per \Cref{prop:approx_error},  the uniform approximation guarantee of RFF holds with respect to a given base measure $\nu$.
    In particular, when only the coefficients $\boldsymbol{\alpha}$ are updated, the resulting finite-dimensional RFF representation can only approximate functions that lie in the function space induced by the kernel associated with $\nu$.
    For the Gaussian kernel, the base measure $\nu$ is determined by the bandwidth parameter $\sigma$. 
    Consequently, the choice of $\sigma$ directly affects the expressiveness of the corresponding function space. 
    However, selecting an appropriate bandwidth \textit{a priori} can be challenging in the presence of time-varying or non-stationary disturbances, where the underlying disturbance characteristics may change over time. 
    This observation motivates the online adaptation of both the kernel coefficients $\boldsymbol{\alpha}$ and the kernel bandwidth $\sigma$.
\end{remark}

\subsection{Policy Optimization with Analytical Gradients}\label{subsec:BPTT}
Consider the quadrotor dynamics in \cref{eq:quadrotor}, the dynamcis is differentiable if the policy $\pi: \mathcal{O} \rightarrow \mathcal{U}$ and the disturbance $\boldsymbol{h}$ is differentiable. If we further consider a differentiable reward function $r: \mathcal{X} \times \mathcal{U} \rightarrow \mathbb{R}$, then all components are fully differentiable
and allow analytic gradient backpropagation through the simulation for policy optimization, \ie Back-Propagation Through Time (BPTT)~\cite{metz2021gradients}.
Specifically, the objective of BPTT for policy optimization is to maximize the cumulative reward over an $N$-step rollout of the policy $\pi_{\boldsymbol\psi}$ parameterized by $\boldsymbol\psi$, \ie
\begin{equation}
    \max R(\boldsymbol{\psi}) = \max_{\boldsymbol{\psi}} \frac{1}{N}\sum_{i=1}^{N} r\left(\boldsymbol{x}_i, \boldsymbol{u}_i\right) = \max \frac{1}{N}\sum_{i=1}^{N} r\left(\boldsymbol{x}_i, \pi_{\boldsymbol{\psi}}(\boldsymbol{o}_i)\right).
\end{equation}
Then $\boldsymbol{\psi}$ can be updated through analytic gradient $\nabla_{\boldsymbol{\psi}}R$ with step size $\eta_{\boldsymbol{\psi}}$, \ie $$\boldsymbol{\psi}_{k+1} = \boldsymbol{\psi}_{k} - \eta_{\boldsymbol{\psi}}\nabla_{\boldsymbol{\psi}_k}R.$$

\subsection{Online Least-Squares Estimation}\label{subsec:OLS}
Based on \Cref{prop:approx_error}, the approximation accuracy depends on the parameter of the base measure $\nu$, \ie $\sigma$, and the parameters $\boldsymbol{\alpha}$. Therefore, we aim to utilize an online least-square estimation algorithm to minimize approximation error.
Given a data point $\left( \boldsymbol{z}_t, \; \boldsymbol{h}\left(\boldsymbol{z}_t\right) \right)$ observed at time $t$, we employ an online least-squares algorithm that updates the parameters $\hat\sigma_t$ and $\hat{\boldsymbol{\alpha}}_t \triangleq \left[ \boldsymbol{\alpha}_{i,t}^\top, \; \dots, \; \boldsymbol{\alpha}_{M,t}^\top\right]^\top$ to minimize the approximation error $l_t = \| \boldsymbol{h}\left(\boldsymbol{z}_t\right) - \hat{\boldsymbol{h}}\left(\boldsymbol{z}_t\right) \|^2$, where $ \hat{\boldsymbol{h}}(\cdot) \triangleq  \frac{1}{M} \sum_{i=1}^{M} \Phi\left(\cdot, \boldsymbol{\theta}_i \right) \hat{\boldsymbol{\alpha}}_{i,t} $ and $\Phi\left(\cdot,\boldsymbol{\theta}_i\right) $ is the random Fourier feature as in \Cref{subsec:RFF}. 
Specifically, we can obtain the gradient $\nabla_{\hat{\sigma}_t} \triangleq \nabla_{\hat{\sigma}_t} l_t$, $\nabla_{\hat{\boldsymbol{\alpha}}_t} \triangleq \nabla_{\hat{\boldsymbol\alpha}_t} l_t$ and update by $\hat{\sigma}_{t+1}^\prime= \hat{\sigma}_t- \eta \nabla_{\hat{\sigma}_t}$, $\hat{\boldsymbol\alpha}_{t+1}^\prime= \hat{\boldsymbol\alpha}_t- \eta \nabla_{\hat{\boldsymbol{\alpha}}_t}.$

The online least-squares estimation algorithm generalizes the seminal online gradient descent algorithms to the non-convex loss function. In the convex case, the online gradient descent algorithm enjoys provable performance guarantee~\cite{hazan2016introduction}. Specifically, the algorithm achieves asymptotically the same cumulative estimation error as the optimal parameters of $\min \sum_{t=1}^T l_t$.

% The above online least-squares estimation algorithm enjoys an $\calO\left(\sqrt{T}\right)$ regret bound, per the regret bound of \OGD~\cite{hazan2016introduction}.

% \begin{proposition}[Regret Bound of Online Least-Squares Estimation~\cite{hazan2016introduction}]\label{theorem:OGD}
%     Assume $\eta=\calO\left({1}/{\sqrt{T}}\right)$.  Then,
%     \begin{equation}
%        \SReg\triangleq \sum_{t=1}^{T} l_t \left(\alpha_t\right) - \sum_{t=1}^{T} l_t \left(\alpha^{\star}\right)  \leq \calO\left(\sqrt{T}\right),
%     \end{equation}
%     where $\alpha^{\star} \triangleq \underset{\alpha \in \calD}{\operatorname{\textit{argmin}}}\;\sum_{t=1}^{T} l_t \left(\alpha\right)$ is the optimal parameter that achieves lowest cumulative loss in hindsight.
% \end{proposition}

% The online least-squares estimation algorithm thus asymptotically achieves the same estimation error 
% as the optimal parameter $\alpha^{\star}$ since $\lim_{T\rightarrow\infty} \;\SReg/T = 0$. 

\subsection{Algorithm}\label{subsec:algo}
% A schematic diagram of the proposed algorithm is given in \Cref{fig:framework}.
The proposed algorithm is composed of two phases: offline learning~(\Cref{alg:BPTT}) and online adaptation~(\Cref{alg:MODA}). 

First, based on the function approximation introduced in \Cref{subsec:RFF}, we can parameterize the dynamics as $\boldsymbol{x}_{t+1} = \boldsymbol{f}(\boldsymbol{x}_{t}) + \boldsymbol{g}(\boldsymbol{x}_{t}) \boldsymbol{u}_{t} + \frac{1}{M} \sum_{i=1}^{M} \Phi\left(\boldsymbol{z}_t, \boldsymbol{\theta}_i \right) {\boldsymbol{\alpha}}_{i,k}$, with unknown parameters $\sigma$ and $\boldsymbol{\alpha}$. Those parameters provide dynamics information that the policy can leverage for online domain adaptation. Therefore, in \Cref{alg:BPTT}, we train an adaptive policy $\pi$ with parameters $\sigma$ and $\boldsymbol{\alpha}$ included in the observations. Since the RFF is differentiable, we leverage BPTT for policy optimization.

In the online phase~(\Cref{alg:MODA}), we collect the data for online adaptation as we apply the policy~(lines 4-5). Then we use online least-square estimation to learn the parameters $\sigma$ and $\boldsymbol{\alpha}$~(lines 6-9). Those learned parameters are fed into the policy observation to enable domain adaptation.

\setlength{\textfloatsep}{-0.1mm}
\begin{algorithm}[t]
% \small
\caption{Policy Optimization (Offline).}
\begin{algorithmic}[1]
    \REQUIRE Number of random Fourier features $M$; random Fourier feature parameters $\{\boldsymbol{\theta}\}_{1,\dots,M}$; domain set $\calD_\sigma$ and $\calD_{\boldsymbol{\alpha}}$;  gradient descent learning rate $\eta_{\boldsymbol{\psi}}$; number of epoch $K$; rollout step $N$.
    \ENSURE Learned policy $\pi_{\boldsymbol{\psi}}$.
    \medskip
        \STATE Formulate $\boldsymbol\Phi\left(\cdot, \boldsymbol\theta_i\right)$, where $i \in \{1, \dots, M\}$;
    \FOR {each epoch $k = 1, \dots, K$}
    \STATE Initialize $\boldsymbol{x}_1$;
    \STATE Reset reward $R(\boldsymbol\psi) = 0$;
    \STATE Randomly sample $\sigma_k$ from $\calD_\sigma$ and $\boldsymbol{\alpha}_k$ from $\calD_{\boldsymbol{\alpha}}$;
    \FOR {each rollout step $t = 1, \dots, N$}
        \STATE Apply control input $\boldsymbol{u}_t = \pi_{\boldsymbol\psi_k}(\boldsymbol{o}_t)$  with ${\sigma}_k$ and ${\boldsymbol\alpha}_{k}$;
        \STATE Receive reward $r_t (\boldsymbol{x}_t, \boldsymbol{u}_t)$;
        \STATE Cumulate $R(\boldsymbol\psi) \leftarrow R(\boldsymbol\psi) + \frac{1}{N}r_t$;
        \STATE Observe state $\boldsymbol{x}_{t+1} = \boldsymbol{f}(\boldsymbol{x}_{t}) + \boldsymbol{g}(\boldsymbol{x}_{t}) \boldsymbol{u}_{t} + \frac{1}{M} \sum_{i=1}^{M} \Phi\left(\boldsymbol{z}_t, \boldsymbol{\theta}_i \right) {\boldsymbol{\alpha}}_{i,k}$;
        \ENDFOR
    \STATE Obtain gradient $\nabla_{\boldsymbol{\psi}}R$;
    \STATE Update $\boldsymbol\psi$ by $\boldsymbol{\psi}_{k+1} = \boldsymbol{\psi}_{k} - \eta_{\boldsymbol{\psi}}\nabla_{\boldsymbol{\psi}_k}R$.
    \ENDFOR
\end{algorithmic}\label{alg:BPTT}
\end{algorithm}

\setlength{\textfloatsep}{-0.1mm}
\begin{algorithm}[t]
% \small
\caption{Model-based Domain Adaptation (Online).}
\begin{algorithmic}[1]
    \REQUIRE Learned policy $\pi_{\boldsymbol{\psi}}$; number of random Fourier features $M$; random Fourier feature parameters $\{\boldsymbol{\theta}\}_{1,\dots,M}$; domain set $\calD_\sigma$ and $\calD_{\boldsymbol{\alpha}}$;  gradient descent learning rate $\eta$.
    \ENSURE At each time step $t=1,\ldots,T$, control input $u_{t}$.
    \medskip
        \STATE Initialize $\boldsymbol{x}_1$, $\hat{\sigma}_1 \in \calD_\sigma$, $\hat{\boldsymbol\alpha}_{i,1} \in \calD_{\boldsymbol{\alpha}}$; 
        \STATE Formulate $\boldsymbol\Phi\left(\cdot, \boldsymbol\theta_i\right)$, where $i \in \{1, \dots, M\}$;
    \FOR {each time step $t = 1, \dots, T$}
    \STATE Apply control input $u_t = \pi_{\boldsymbol{\psi}}(\boldsymbol{o}_t)$  with $\hat{\sigma}_t$ and $\hat{\boldsymbol\alpha}_{t}$;
        \STATE Observe state $\boldsymbol{x}_{t+1}$, and calculate disturbance via $\boldsymbol{h}(\boldsymbol{z}_t) = \boldsymbol{x}_{t+1} - \boldsymbol{f}(\boldsymbol{x}_{t}) - \boldsymbol{g}(\boldsymbol{x}_{t}) \boldsymbol{u}_{t}$;
        \STATE Formulate estimation loss $l_t\left(\hat{\sigma}_t,\;\hat{\boldsymbol{\alpha}}_t\right) \triangleq \left\| \boldsymbol{h}\left(\boldsymbol{z}_t\right) -  \frac{1}{M} \sum_{i=1}^{M} \Phi\left(\boldsymbol{z}_t, \boldsymbol{\theta}_i \right) \hat{\boldsymbol{\alpha}}_{i,t}  \right\|^2$;
        \STATE Calculate gradient $\nabla_{\hat{\sigma}_t}$ and $\nabla_{\hat{\boldsymbol{\alpha}}_t}$;
        \STATE Update $\hat{\sigma}_{t+1}^\prime= \hat{\sigma}_t- \eta \nabla_{\hat{\sigma}_t}$ and
                $\hat{\boldsymbol\alpha}_{t+1}^\prime= \hat{\boldsymbol\alpha}_t- \eta \nabla_{\hat{\boldsymbol{\alpha}}_t}$;
        \STATE Project $\hat{\sigma}_{t+1}^\prime$ onto $\calD_\sigma$ and each $\hat{\boldsymbol\alpha}_{i,t+1}^\prime$ onto $\calD_{\boldsymbol{\alpha}}$;
        \ENDFOR
\end{algorithmic}\label{alg:MODA}
\end{algorithm}

	\section{Experiments}
\label{sec:conclusion}
In this section, we present our experimental results. We first introduce our training precedure and algorithm parameters in \Cref{subsec:train}, and benchmark algorithms in \Cref{subsec:benchmark}. We then present our simulation results in \Cref{subsec:sim} and hardware restuls using Crazyflies in \Cref{subsec:hw}. 

\subsection{Policy Training and Algorithm Parameters}\label{subsec:train}
For the policy network, we use a 3-layer MLP with a hidden dimension of 512. The policy observation $\boldsymbol{o}_t$ has a size of $187$, including body-frame velocity $\boldsymbol{R}^\top\boldsymbol{v}_t$, quaternion $\boldsymbol{q}_t$, thrust $T_{t-1}$, angular velocity command $\boldsymbol{\omega}_{t-1}$, 10 relative future reference trajectory point $\Delta\boldsymbol{x}$ over a time horizon $1.0s$, and parameters $\boldsymbol{\alpha}_t$ and $\sigma_t$. The policy outputs the mass-normalized thrust $T_{t}/m$ and angular velocity command $\boldsymbol{\omega}_{t}$. 

For BPTT, we trian the policy to track a lemniscate trajectory with 500 environments and 300 epochs using JAX. Each episode has a maximum of 250 simulation steps with a time step of $0.02s$. We use an initial step size of $0.001$ with a decay following a cosine function\footnote{\url{https://optax.readthedocs.io/en/latest/api/optimizer_schedules.html}}. For each episode, we sample parameters $\boldsymbol{\alpha}$ from a Gaussian distribution $\mathcal{N}(\boldsymbol{0}, \frac{1}{2}\boldsymbol{I})$ and $\sigma$ from an uniform distribution $[0.001, \; 1]$.
We use Gaussian Kernel and $M=25$ random features, which leads to $\boldsymbol{\alpha}\in\mathbb{R}^{75}$. For reward, we use as reward the negative tracking errors in position, quaternion, velocity, and control deviation from hovering command, with weights $2.5$, $0.5$, $0.1$, and $0.01$ for simulation, and weights $2.5$, $0.5$, $0.1$, and $0.1$ for hardware.
The training takes $50s$ on a RTX 4090 GPU.

For the online update of $\boldsymbol{\alpha}$ and $\sigma$, we use step size of $\eta=0.1$ for simulation and $\eta=0.02$ for hardware. We initialize $\boldsymbol{\alpha}$ as zeros and $\sigma=0.5$. We use as features $\boldsymbol{z}$ the body-frame velocity $\boldsymbol{R}^\top\boldsymbol{v}_t$, quaternion $\boldsymbol{q}_t$, normalized thrust $T_{t-1}/mg$, angular velocity command $\boldsymbol{\omega}_{t-1}$.

Compared to simulation, we use higher penalty on control action and lower learning rate of online adaptation to avoid aggressive behaviors in hardware.

\subsection{Benchmark Algorithms}\label{subsec:benchmark}
We compare our methods with both model-free and model-based adaptive control algorithms: DATT~(L1-E)~\cite{huang2023datt}, DATT~(RMA)~\cite{huang2023datt}, MPC~(L1-E)~\cite{huang2023datt}, and MPC~(RFF)~\cite{zhou2025simultaneous}.

Specifically, both DATT~(L1-E) and DATT~(RMA) learn an adaptive control policy via PPO~\cite{schulman2017proximal} conditioned on a vector-valued estimate of the disturbance. DATT~(L1-E) utilizes a state predictor to estimate the disturbance as in L1 adaptive control~\cite{wu2023mathcal} and DATT~(RMA) based on Rapid Motor Adaptation (RMA)~\cite{kumar2021rma}. The RMA is a neural-network module trained by supervised learning that predicts disturbance using a history of state-action pairs.

MPC~(L1-E) and MPC (RFF) use model predictive control with quadrotor dynamics in \cref{eq:quadrotor}. MPC~(L1-E) utilizes a state predictor to estimate the disturbance based on L1 adaptive control~\cite{wu2023mathcal} and augments this estimation to the nominal quadrotor dynamics.
MPC (RFF) parametrizes the disturbance using RFF and updates $\boldsymbol\alpha$ only.
A detailed implementation of the benchmark algorithms is provided in Appendix~\ref{sec:implementation}.

We also compare our method with variants without the $\sigma$ and/or $\boldsymbol{\alpha}$ updates in the ablation study.

\subsection{Numerical Simulations}\label{subsec:sim}
We first conduct numerical simulations using a realistic simulator with a Blade Element Momentum (BEM)~\cite{pan2026learning} model for aerodynamic effects and high-frequency simulation of low-level controller dynamics~\cite{heeg2025learning}.
We simulate for $10s$.
We test the algorithms under various disturbances, including: 
\begin{enumerate}[label=(\roman*)]
    \item aerodynamics effects only; 
    % \item aerodynamics effects plus constant disturbance $[1.0,\;1.0,\;1.0]^\top$;
    \item aerodynamics effects plus sinusoidal disturbance $\frac{1}{2}[\sin \frac{2\pi t}{10},\;\sin \frac{2\pi t}{10},\;\sin \frac{2\pi t}{10}]^\top$; 
\item aerodynamics effects plus switching distrubance from sinusoidal disturbance $\frac{1}{2}[\sin \frac{\pi t}{10},\;\sin \frac{\pi t}{10},\;\sin \frac{\pi t}{10}]^\top$ (from $t=0s$ to $5s$) to constant disturbance $[0.5,\;0.5,\;0.5]^\top$ (from $t=5s$ to $10s$); 
    \item aerodynamics effects plus quadratic-phase sinusoidal disturbance $\frac{1}{2}[\sin \frac{\pi t^2}{10},\;\sin \frac{\pi t^2}{10},\;\sin \frac{ \pi t^2}{10}]^\top$.
\end{enumerate}
% (i) aerodynamics effects only; 
% (ii) aerodynamics effects plus constant disturbance $[1.0,\;1.0,\;1.0]^\top$;
% (iii) aerodynamics effects plus sinusoidal disturbance $\frac{1}{2}[\sin \frac{2\pi t}{10},\;\sin \frac{2\pi t}{10},\;\sin \frac{2\pi t}{10}]^\top$; 
% (iv) aerodynamics effects plus switching distrubance from sinusoidal disturbance $\frac{1}{2}[\sin \frac{2\pi t}{10},\;\sin \frac{2\pi t}{10},\;\sin \frac{2\pi t}{10}]^\top$ (from $t=0s$ to $5s$) to constant disturbance $[0.5,\;0.5,\;0.5]^\top$ (from $t=5s$ to $10s$); 
% (v) aerodynamics effects plus quadratic-phase sinusoidal disturbance $\frac{1}{2}[\sin \frac{\pi t^2}{10},\;\sin \frac{\pi t^2}{10},\;\sin \frac{ \pi t^2}{10}]^\top$.
We use the RMSE in position as the performance metric. 

The results are given in \Cref{table:sim-pos} and Appendix~\ref{sec:app-sim}. In \Cref{table:sim-pos}, our method consistently has the lowest tracking error across all tested scenarios. 
% Specifically, our method improves up to $60\%$ over DATT~(L1-E), $70\%$ over DATT~(RMA), $70\%$ over MPC~(L1-E), and $20\%$ over MPC~(RFF). 
Specifically, our method improves up to $70\%$ over DATT~(L1-E), $70\%$ over DATT~(RMA), $60\%$ over MPC~(L1-E), and $30\%$ over MPC~(RFF). 
The performance is connected to the learning accuracy of RMA, L1-estimator, and RFF. In Appendix~\ref{sec:app-sim}, we show see that RMA results in the largest prediction error, as it highly depends on the training dataset, as pointed out in \cite{huang2023datt}, and RFF has better performance than the L1-estimator.

In addition, DATT variants require the values of disturbance lie in the range of $[-2, \; 2] m/s^2$ during training. This range cannot cover complicated disturbance functions such as switching and quadratic-phase sinusoidal, shown in Appendix~\ref{sec:app-sim}.
While for RFF, this disturbance can still be represented correctly within the training range of $\boldsymbol{\alpha} \sim \mathcal{N}(\boldsymbol{0}, \frac{1}{2}\boldsymbol{I})$ and $\sigma \in [0.001, \; 1]$, shown in Appendix~\ref{sec:app-sim}, providing a better representation of learning a disturbance model.

The last three columns in \Cref{table:sim-pos} present the results of the ablation study. 
% Removing both the $\sigma$ and $\boldsymbol{\alpha}$ updates causes the system to degrade massively (\eg the tracking errors degrade from $2.5~cm$ to over $21~cm$). Removing the $\sigma$ update still results in $10\%$ to $20\%$ performance drop across all scenarios compared to the full pipeline with $\sigma$ and $\boldsymbol{\alpha}$ updates.
Removing both the $\sigma$ and $\boldsymbol{\alpha}$ updates causes the system to degrade massively (\eg the tracking errors degrade from around $4~cm$ to around $40~cm$). Removing the $\sigma$ update still results in $10\%$ to $30\%$ performance drop across all scenarios compared to the full pipeline with $\sigma$ and $\boldsymbol{\alpha}$ updates.

\begin{table*}[h]
    \captionsetup{font=small}
    \centering
    \caption{\textbf{Performance Comparison for the Numerical Simulations in \Cref{subsec:sim}.} The table reports the average value of RMSE in position ($cm$). The \blue{blue} numbers correspond to the \blue{better performance}. Our method achieves better tracking performance than all benchmark algorithms.}
    \label{table:sim-pos}
    \resizebox{\columnwidth}{!}{
    \begin{tabular}{cccccccc}
    \toprule
    Disturbance   & DATT (L1-E) & DATT (RMA) & MPC (L1-E) & MPC (RFF) & Ours (w/o $\sigma$ \& $\boldsymbol{\alpha}$ update)  & Ours (w/o $\sigma$ update) & Ours \cr
    \midrule
     Aerodynamic & $10.54$ & $13.98$ & $7.23$ & $3.49$ & $37.93$ & $3.49$ &  \blue{$3.17$}   \cr
     Aerodynamic + Sinusoidal & $10.50$ & $14.73$ & $9.15$ & $4.44$ & $42.86$ & $5.37$ &    \blue{$4.10$}     \cr
     Aerodynamic + Switching & $10.71$ & $14.22$ & $12.16$ & $5.44$ & $36.52$ & $5.66$ &  \blue{$4.11$}    \cr
     Aerodynamic + Quadratic-Phase Sinusoidal & $10.02$ & $12.75$ & $9.90$ & $5.94$ & $37.22$ & $5.58$  &  \blue{$4.00$}    \cr
    \bottomrule
    \end{tabular}
    }
\end{table*}

\subsection{Crazyflies Experiments}\label{subsec:hw}
We deploy our algorithm on Crazyflie drone~(\Cref{fig:hw}).\footnote{\url{https://www.bitcraze.io/products/old-products/crazyflie-2-1/}}
We test the algorithms under various time-varying disturbances, including: 
% (i) ground effect, (iii) wind disturbance, and (iii) $6g$ payload.
\begin{enumerate}[label=(\roman*)]
    \item ground effect, where part of the trajectory is $7cm$ above a box and part of the trajectory is $72cm$ above the ground, creating a switching between w/ and w/o ground effect; 
    \item time-varying wind disturbance, where the wind speed varies from $1.5m/s$ to $2.5m/s$;
    \item $6g$ payload suspended to the bottom of the drone, which weighs $20\%$ of the drone mass.
\end{enumerate}

We use the RMSE in position as the performance metric. 

The results are given in \Cref{table:hw-pos} and Appendix~\ref{sec:app-hw}.
In \Cref{table:hw-pos}, our method consistently has the lowest tracking error across all tested scenarios. 
% Specifically, our method improves up to $10\%$ over DATT~(L1-E), $20\%$ over DATT~(RMA), $30\%$ over MPC~(L1-E), and $20\%$ over MPC~(RFF).
Specifically, our method improves up to $10\%$ over DATT~(L1-E), $25\%$ over DATT~(RMA), $34\%$ over MPC~(L1-E), and $20\%$ over MPC~(RFF).

The last three columns in \Cref{table:hw-pos} present the results of the ablation study. 
% Removing both the $\sigma$ and $\boldsymbol{\alpha}$ updates causes the system to degrade from $4-5~cm$ to over $8-9~cm$). 
Removing both the $\sigma$ and $\boldsymbol{\alpha}$ updates causes the system to degrade from $6-9~cm$ to over $10-15~cm$). 
Removing the $\sigma$ update still results in $3\%$ to $10\%$ performance drop across all scenarios compared to the full pipeline with $\sigma$ and $\boldsymbol{\alpha}$ updates. Compared to the simulation results in \Cref{subsec:sim}, the performance of these variants is closer, since we use a smaller step size $\eta=0.02$ to avoid aggressive updates with noisy real-world data.

% RMSE
\begin{table*}[h]
    \captionsetup{font=small}
    \centering
    \caption{\textbf{Performance Comparison for the Hardware Experiments in \Cref{subsec:hw}.} The table reports the average value of RMSE in position ($cm$). The \blue{blue} numbers correspond to the \blue{better performance}. Our method achieves better tracking performance than all benchmark algorithms.}
    \label{table:hw-pos}
    \resizebox{\columnwidth}{!}{
    \begin{tabular}{cccccccc}
    \toprule
     Disturbance & DATT (L1-E) & DATT (RMA) & MPC (L1-E) & MPC (RFF) & Ours (w/o $\sigma$ \& $\boldsymbol{\alpha}$ update)  & Ours (w/o $\sigma$ update) & Ours \cr
    \midrule
     Ground & $7.12$  & $8.64$ & $9.72$ & $8.25$ & $14.41$ & $6.92$ &   \blue{$6.41$}    \cr
     Wind & $9.95$  & $9.39$ & $10.99$ & $9.91$ & $10.78$ & $9.76$ &   \blue{$8.80$}    \cr
     Payload & $7.37$  & $7.20$ & $9.73$ & $7.21$ & $15.20$ & $6.95$ &  \blue{$6.68$}    \cr
    \bottomrule
    \end{tabular}
    }
\end{table*}

	\section{Limitations}
\label{sec:limitations}
The training distribution of $\boldsymbol{\alpha}\sim\mathcal{N}(\boldsymbol{0},\;\frac{1}{2}\boldsymbol{I})$ and $\sigma \in [0.001, \; 1]$ provides a rich representation for learning disturbance, such as aerodynamics, payload, wind, and ground effects. Handling out-of-distribution values of $\boldsymbol{\alpha}$ and $\sigma$ encountered during online adaptation is not considered. An OOD-detection module based on learned $\boldsymbol{\alpha}$ and $\sigma$ values and rapid online policy re-training mechanism~\cite{pan2026learning} can be incorporated into the pipeline.
We demonstrate the capability of our method under non-stationary disturbance in practice, while formal stability guarantees are not considered in this paper. 
Though not observed in our experiments, the proposed approach may suffer from bursting behavior, as commonly encountered in adaptive control methods, during rapid distribution shifts or poorly conditioned online adaptation~\cite{anderson1985adaptive}. Incorporating an active learning mechanism to improve persistent excitation for disturbance learning may help mitigate such behavior.

\section{Conclusion}
\label{sec:conclusion}
We presented an algorithm for efficient domain-adaptive policy learning via kernel representations. We modeled the unknown disturbance with a differentiable kernel approximation based on random Fourier features, in which both the kernel coefficients and the bandwidth are learnable. 
In the offline phase, we randomly sampled the coefficients and bandwidth to generate a diverse family of disturbance profiles and trained the control policy via differentiable simulation with analytical gradients, taking only $50$ seconds on a single RTX 4090 GPU. 
In the online phase, we deployed the policy zero-shot and updated both the coefficients and the bandwidth jointly at every control step via online least-squares estimation, allowing the basis itself to adapt to the disturbance encountered at deployment. 
We validated the algorithm in high-fidelity numerical simulations under aerodynamic effects, sinusoidal, switching, and quadratic-phase sinusoidal disturbances, and on Crazyflie hardware under suspended payload weighing 20\% of the drone mass, on-and-off-switching ground effects, and time-varying wind. Across all settings, our method outperformed state-of-the-art learning-based and model-based adaptive control baselines, improving position tracking by up to $60 - 70\%$ in simulation and up to $30\%$ on hardware. 
Future work includes incorporating out-of-distribution detection and rapid online re-training to handle disturbances outside the training range of the kernel parameters, establishing formal stability guarantees under non-stationary disturbances, and exploring active learning mechanisms to ensure persistent excitation during online adaptation.

	\clearpage
	% The acknowledgments are automatically included only in the final and preprint versions of the paper.
    \acknowledgments{This work was supported in part by the National Science Foundation
	(NSF) CAREER Award No. 2337412, and the Army Research Office (ARO)
	Early Career Program Award W911NF-25-1-0280.}
	
	%===============================================================================

	\bibliography{reference}  % .bib

	% \newpage
\begin{appendices}

% \newpage
\section{Related Works}
\label{sec:related}

We discuss work on robust and adaptive control, learning-based control, and online learning for control.

\paragraph*{Robust and adaptive control.}
Robust control algorithms select control inputs assuming a worst-case realization of disturbances~\cite{zhou1998essentials,mayne2005robust,zhou2023safe,martin2024guarantees,liu2024robust}, which can be conservative. Adaptive control methods often assume parametric uncertainty additive to the known system dynamics, \eg parametric uncertainty in the form of unknown coefficients multiplying known basis functions~\cite{slotine1991applied,krstic1995nonlinear,ioannou1996robust}. These coefficients are updated online and generate an adaptive control input to compensate for the estimated disturbances. However, a fixed set of basis functions can be difficult to generalize across different scenarios, where the disturbance may lie beyond the span of the given basis functions. \cite{tal2020accurate,wu2023mathcal,tao2024robust,das2024robust,hanover2021performance} directly estimate the value of unknown disturbances. However, the methods therein, as well as the relevant methods in~\cite{boffi2021regret,boffi2022nonparametric,jia2023evolver}, focus on adaptive control to compensate for the estimated disturbances, instead of leveraging a model of the disturbances for learning control policy.

\paragraph*{Learning-based control.}
\cite{sanchez2018real,tobin2017domain,ramos2019bayessim,lee2020learning,li2025reinforcement,kumar2021rma,xue2026robust,zhang2025learning,huang2023datt} train a neural-network policy using training data collected offline. \cite{tobin2017domain,ramos2019bayessim,lee2020learning,li2025reinforcement} train the policy with data collected from random environments such that the controller can exhibit robustness to unseen environments~\cite{mohri2018foundations,abu2012learning}. \cite{kumar2021rma,xue2026robust,zhang2025learning,huang2023datt} train the policy conditioned on the output of an environment encoder to achieve domain adaptation~\cite{farahani2021brief,ben2006analysis}. Our approach falls into the category of domain adaptation. Compared to \cite{kumar2021rma,xue2026robust,zhang2025learning,huang2023datt}, we use random Fourier features to generate a diverse disturbance profile for offline training and to approximate an unknown disturbance model online that provides environmental information to the policy for online adaptation. Both the kernel coefficients and the kernel bandwidth are learnable, which enables representation of disturbances not tied to \textit{a priori} chosen basis functions. By leveraging differentiable simulation, we train a policy that can be deployed zero-shot to the real world with only $50$ seconds of training time on an RTX 4090 GPU. \cite{wu2023daydreamer,nan2025efficient,pan2026learning} directly learn the policy online on the hardware, but this can take from seconds to hours to finish the training. In contrast, we learn the adaptive policy offline and achieve zero-shot transfer to hardware and real-time online adaptation.

\paragraph*{Online learning for control.}
Online learning algorithms select control inputs based on past information only since they assume no model that can be used to simulate the future evolution of the disturbances~\cite{hazan2022introduction,agarwal2019online,zhao2022non,zhou2023safecdc,zhou2023saferal,zhou2023efficient}. These methods often quantify the performance of the algorithms through \textit{regret}, \ie the suboptimality against an optimal clairvoyant controller that knows the unknown disturbances and dynamics, and they make no parametric assumption on how the disturbance evolves over time. However, in practice, they have been observed to be sensitive to the choice of the tuning parameters~\cite{zhou2023safecdc}. \cite{zhou2025simultaneous,tsiamis2024predictive,zhou2025adaptive} overcome the sensitivity by learning a model for predictive control. However, similar to adaptive control with fixed basis functions, they lack adaptability to learning disturbances beyond the reach of the given basis functions. We online update the kernel bandwidth of the basis functions, enabling automatic basis function learning for different disturbances.

\section{Benchmark Algorithm Implementation}\label{sec:implementation}

\subsection{DATT (L1-E)}

DATT (L1-E) is implemented based on the official DATT repository in~\cite{huang2023datt}. We use the DATT trajectory-tracking policy with wind-adaptive configuration and enable the online L1-estimation module during training in simulation. By default, the range of disturbance in each axis is $[-2, \; 2] \; m/s^2$.
The policy observation has a total dimension of $46$, consisting of the body-frame position $\boldsymbol{R}_t^\top \boldsymbol{p}_t$, the body-frame velocity $\boldsymbol{R}_t^\top \boldsymbol{v}_t$, the quaternion attitude $\boldsymbol{q}_t$, the estimation of disturbance by L1-estimator, the current body-frame position error, and $10$ future body-frame reference position-error waypoints. Specifically, the body-frame position, the body-frame velocity, the estimation of disturbance, and the current position error each has dimension of $3$, and the quaternion has dimension $4$. The future reference sequence contributes a total dimension of $30$. 
The L1-estimator uses adaptation matrix parameter $A=-0.01$, update time step $0.02$\,s, and low-pass filter coefficient $0.99$. The disturbance estimate is updated online at every control step and fed into the policy observation.

\subsection{DATT (RMA)}
DATT (RMA) is also based on the official DATT repository~\cite{huang2023datt}. It uses the same DATT (L1-E) training setup. The difference is that the estimation of disturbance by a RMA module.
The RMA module is a neural-network that uses a history of state and action pairs with length of $50$ steps. Each history element contains position, velocity, quaternion, and previous action information, resulting in a $14$-dimensional observation vector for each step. The network consists of three one-dimensional convolutional layers with $64$ channels, followed by three fully connected layers with output dimensions of $32$, $32$, and $3$, respectively. The final output is a $3$-dimensional vector of the estimated disturbances.

\subsection{MPC for Numerical Simulations}
MPC (L1-E) is implemented using the sampling-based MPC controller from~\cite{huang2023datt}, together with the same online L1-estimator module used in DATT (L1-E).
The sampling-based MPC state includes position, velocity, quaternion attitude, and additional control variables required by the rollout dynamics. The controller optimizes over collective thrust and body-rate commands. The planning horizon is $H=50$ steps, corresponding to $1.0$\,s with sampling time $0.02s$. At each control step, sampling-based MPC evaluates $8192$ candidate control sequences. The temperature parameter is set to $0.1$.
The control perturbation standard deviation is set to $[0.25, 1, 1, 1]$ for weight-normalized thrust and body-rate commands. Roll and pitch rates are clipped to $[-6,6]$\,rad/s, and yaw rate is clipped to $[-4,4]$\,rad/s. The rollout dynamics include first-order thrust and body-rate lag with a coefficient of $0.4$.
The quadratic cost includes position tracking, attitude tracking, velocity tracking, and control deviation from the hovering command. The corresponding weights are $2.5$, $0.5$, $0.1$, and $0.01$. 

MPC (RFF) uses the sampl-based MPC implemented in~\cite{huang2023datt} with the online Random Fourier Feature residual learning module from~\cite{zhou2025simultaneous}.
The MPC parameters are kept the same as in MPC (L1-E).
The RFF residual model uses an input vector with a total dimension of $11$, consisting of the quaternion attitude, body-frame velocity, normalized collective thrust, and body-rate command. 
The model employs $25$ random Fourier features. The random feature matrix is initialized from a Gaussian distribution with standard deviation of $\sigma=0.5$, while the phase offsets are initialized uniformly over the interval $[0, 2\pi]$. The coefficient $\boldsymbol{\alpha}$ has shape $25 \times 3$ and is initialized to zero.
Online adaptation of the coefficients $\boldsymbol{\alpha}$ is performed at every control step using a learning rate of $0.1$, while $\sigma$ is fixed. The learned residual model is augmented to the nominal model for MPC rollout.

\subsection{MPC for Hardware Experiments}
We implement optimization-based MPC methods for the hardware experiments, based on~\cite{LlanesICRA2024}. The quadratic cost has weights of $20$, $10$, $2$, and $1$ for position tracking, attitude tracking, velocity tracking, and control deviation from the hovering command. For MPC (L1-E), the L1-estimator is kept the same as in the sampling-based MPC. For MPC (RFF), the learning rate of $\boldsymbol{\alpha}$ is $0.02$.

\section{Results of Numerical Simulations}\label{sec:app-sim}

\begin{figure}[h]
    \centering
    \includegraphics[width=0.6\textwidth]{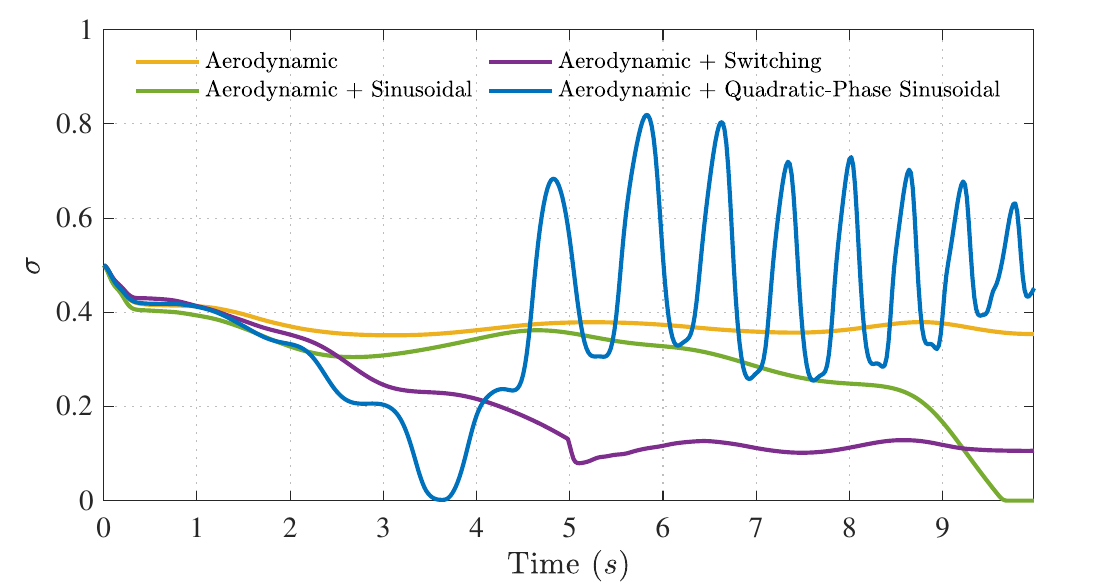}
    \caption{\textbf{Evolution of $\sigma$ Under Different Disturbance in \Cref{subsec:sim}.} The value of ${\sigma}$ lie within the range of $[0.001, \; 1]$ used in training.}
    \label{fig:sigma}
    \vspace{-5mm}
\end{figure}

\begin{figure}[h]
    \centering
    % \hspace*{-0.05\textwidth}
    \includegraphics[width=1.\textwidth]{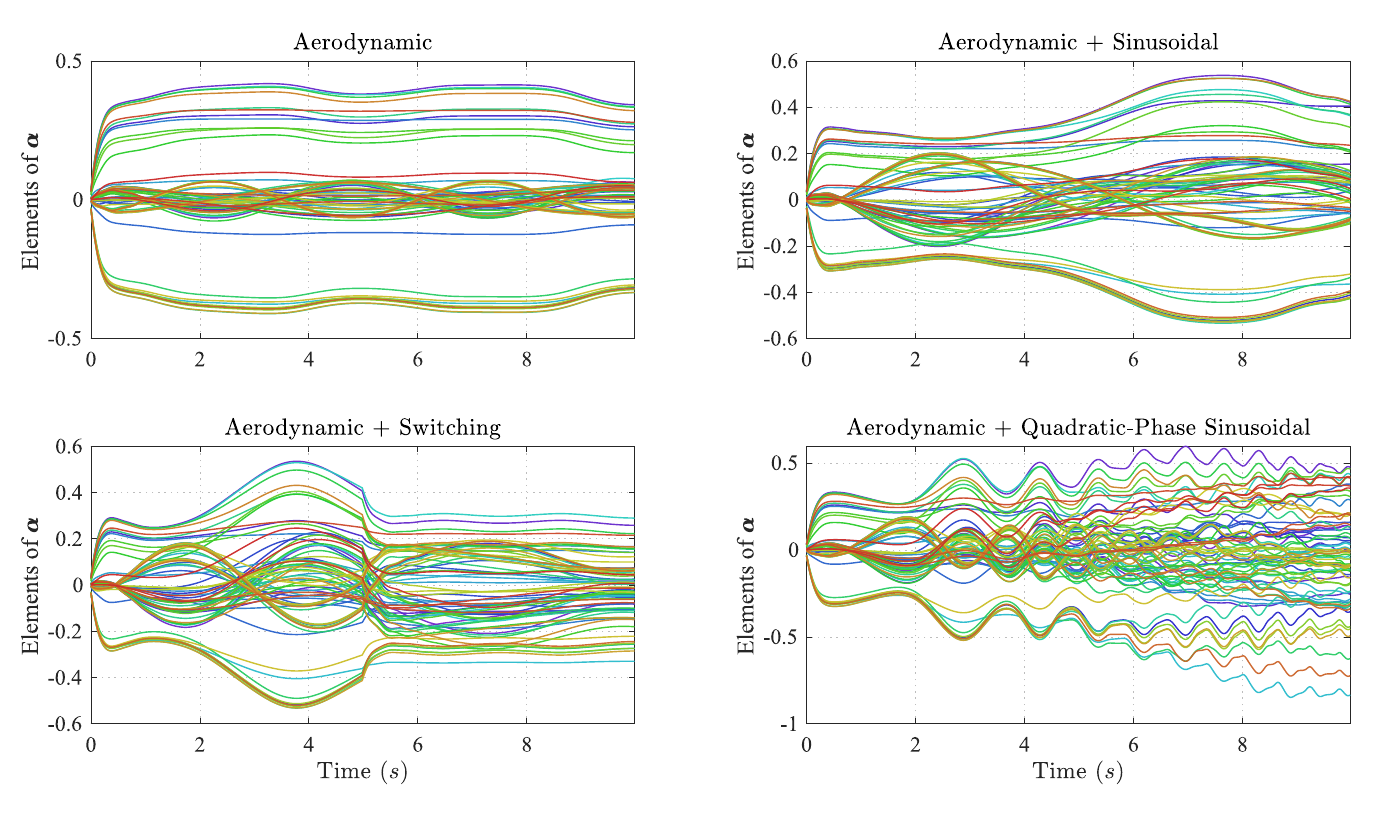} 
    \caption{\textbf{Evolution of $\boldsymbol{\alpha}$ Under Different Disturbance in \Cref{subsec:sim}.} All 75 elements of $\boldsymbol{\alpha}$ are shown and each curve represents an averaged element of $\boldsymbol{\alpha}$. All elements of $\boldsymbol{\alpha}$ lie within 3 times standard deviation of the Gaussian distribution $\mathcal{N}(\boldsymbol{0}, \frac{1}{2}\boldsymbol{I})$ used in training.}
    \label{fig:alpha}
    \vspace{-5mm}
\end{figure}

\begin{figure}[!]
    \centering
    \includegraphics[width=.8\textwidth]{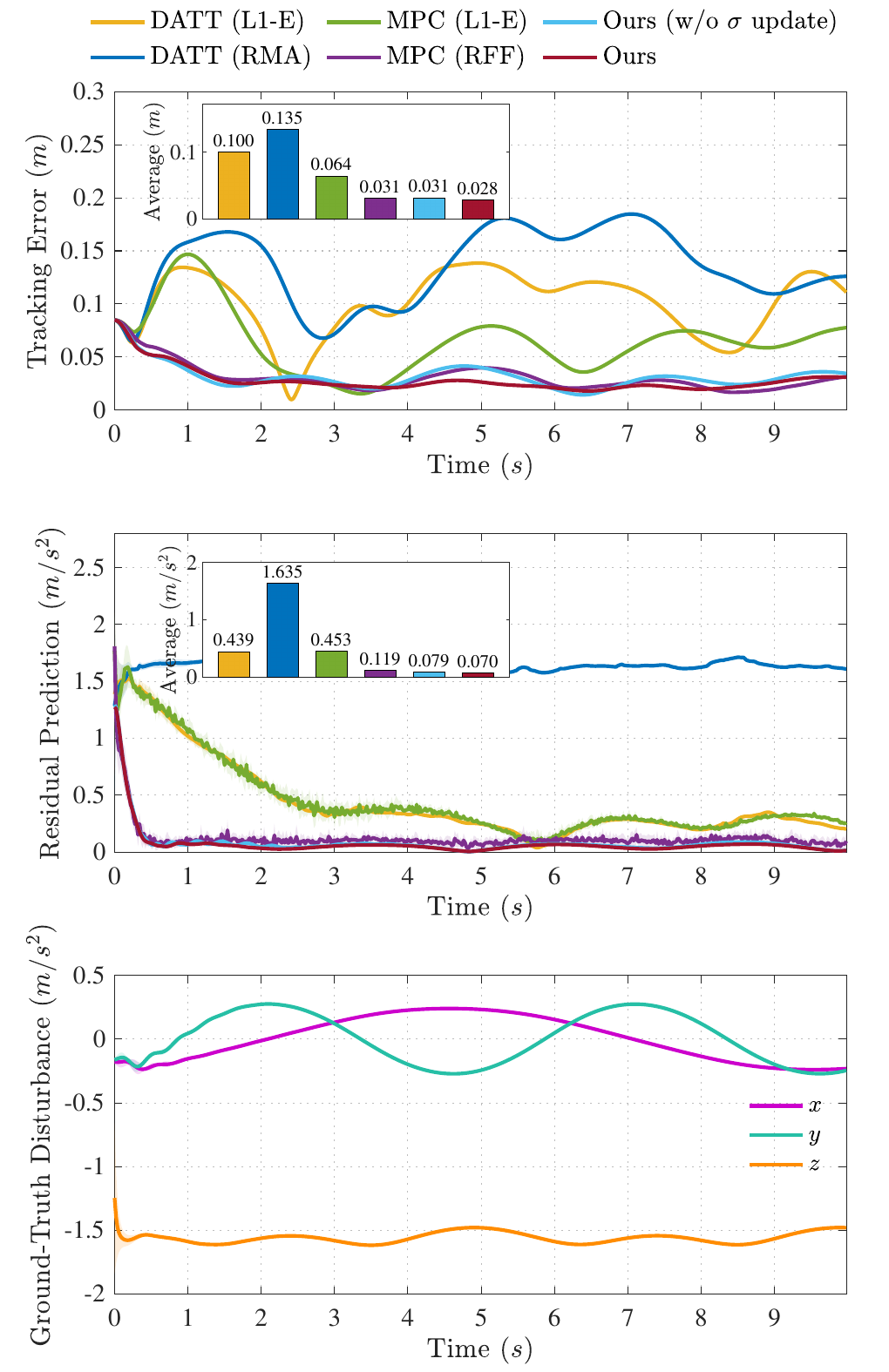}
    \caption{\textbf{Results of Aerodynamic Secnario in the Numerical Simulations in \Cref{subsec:sim}}. Top: Tracking error~($m$), with the inserted plot showing RMSE~($m$). Middle: Norm of residual prediction error~($m/s^2$), with the inserted plot showing the average error~($m/s^2$). Bottom: Ground-truth disturbances~($m/s^2$). }
    \label{fig:sim-drag}
\end{figure}

\begin{figure}[!]
    \centering
    \includegraphics[width=.8\textwidth]{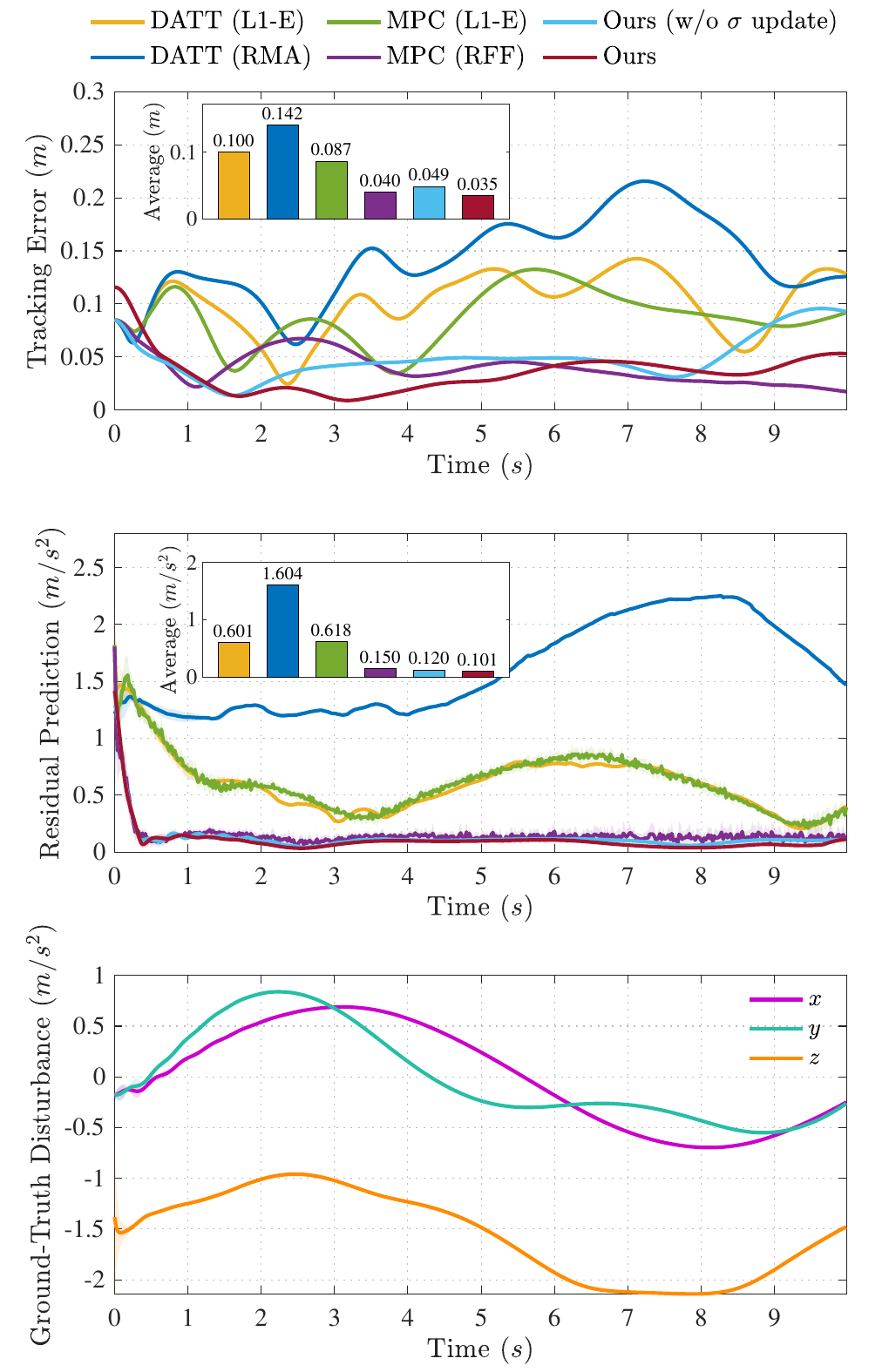}
    \caption{\textbf{Results of Aerodynamic + Sinusoidal Secnario in the Numerical Simulations in \Cref{subsec:sim}}. Top: Tracking error~($m$), with the inserted plot showing RMSE~($m$). Middle: Norm of residual prediction error~($m/s^2$), with the inserted plot showing the average error~($m/s^2$). Bottom: Ground-truth disturbances~($m/s^2$). }
    \label{fig:sim-sin}
\end{figure}

\begin{figure}[!]
    \centering
    \includegraphics[width=.8\textwidth]{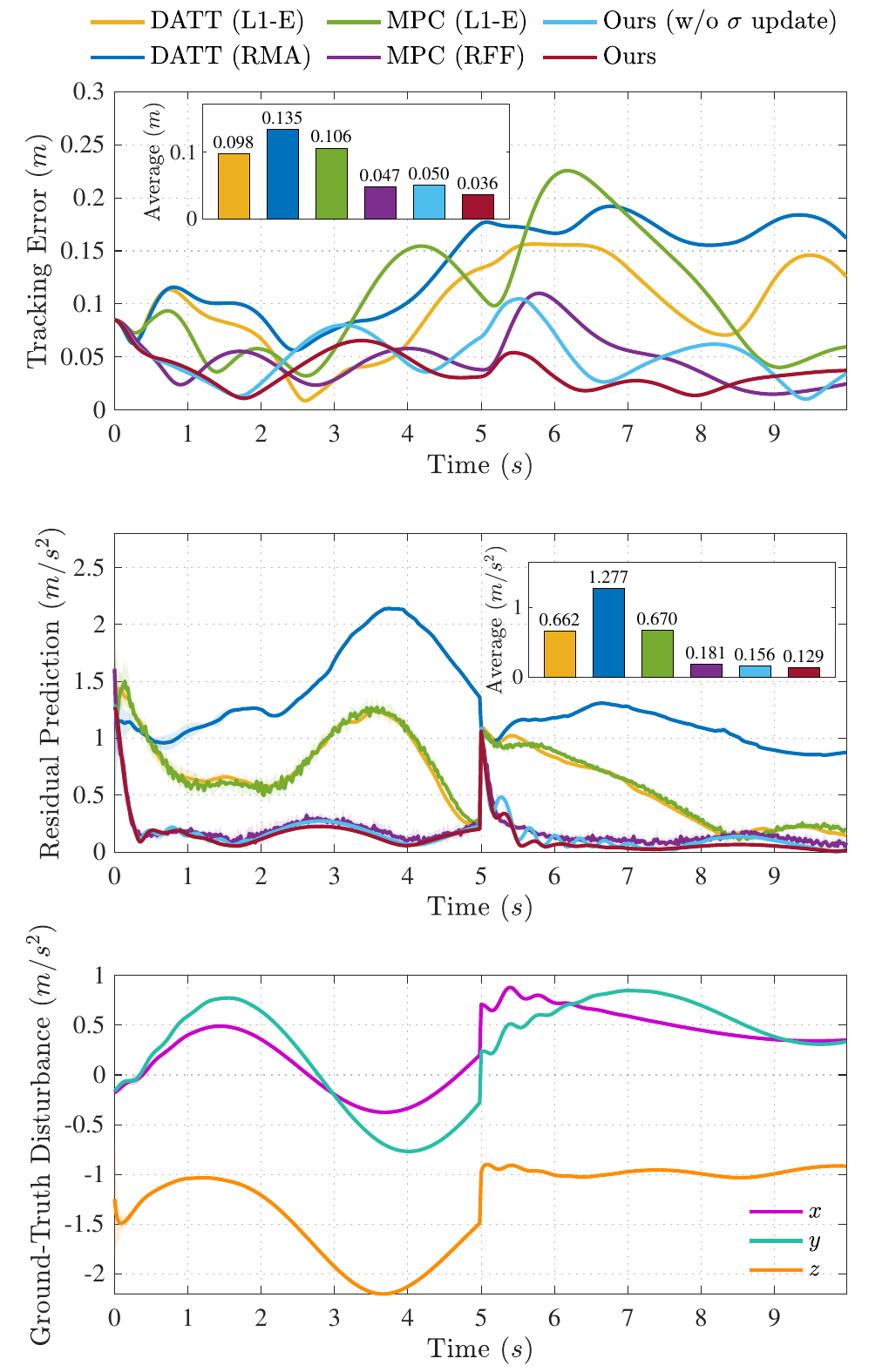}
    \caption{\textbf{Results of Aerodynamic + Switching Secnario in the Numerical Simulations in \Cref{subsec:sim}}. Top: Tracking error~($m$), with the inserted plot showing RMSE~($m$). Middle: Norm of residual prediction error~($m/s^2$), with the inserted plot showing the average error~($m/s^2$). Bottom: Ground-truth disturbances~($m/s^2$). }
    \label{fig:sim-switch}
\end{figure}

\begin{figure}[!]
    \centering
    \includegraphics[width=.8\textwidth]{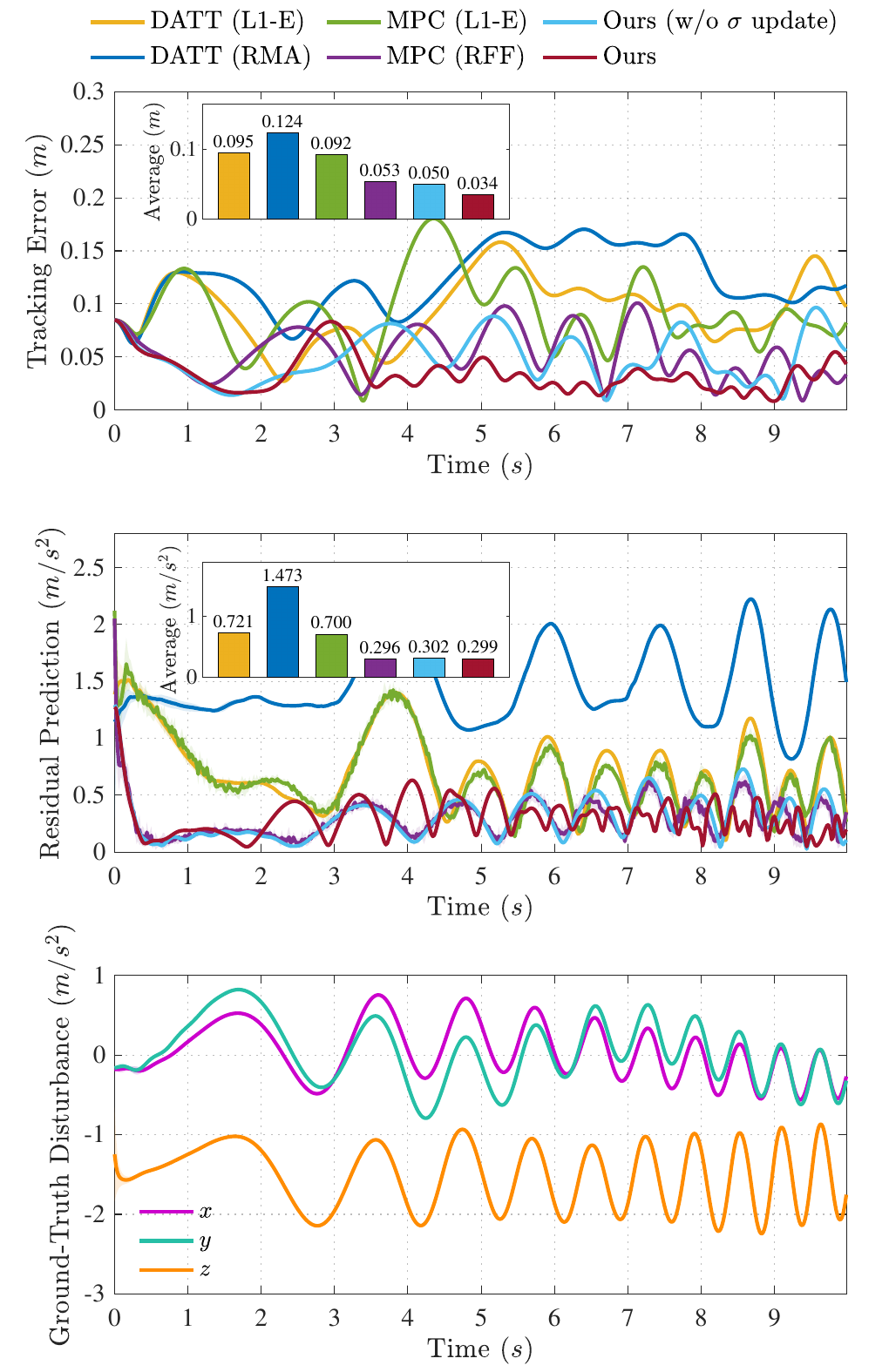}
    \caption{\textbf{Results of Aerodynamic + Quadratic-Phase Sinusoidal Secnario in the Numerical Simulations in \Cref{subsec:sim}}. Top: Tracking error~($m$), with the inserted plot showing RMSE~($m$). Middle: Norm of residual prediction error~($m/s^2$), with the inserted plot showing the average error~($m/s^2$). Bottom: Ground-truth disturbances~($m/s^2$). }
    \label{fig:sim-quad}
\end{figure}

We present results for the numerical simulations in \Cref{subsec:sim}. 
We present the $\boldsymbol\alpha$ and $\sigma$ values in \Cref{fig:sigma} and \Cref{fig:alpha}.
In Figs.~\ref{fig:sim-drag}~-~\ref{fig:sim-quad}, we present (i)  tracking error~$m$, with the inserted plot showing RMSE~($m$);
(ii) norm of residual prediction error~($m/s^2$), with the inserted plot showing the average error~($m/s^2$);
(iii) Ground-truth disturbances~($m/s^2$). 

The plots of tracking errors show that our method achieves the best tracking performance across different disturbances.
The prediction error results illustrate the learning accuracy of RMA, L1-estimator, and RFF. We can see that RMA results in the largest prediction error, as it highly depends on the training dataset as pointed out in \cite{huang2023datt}. RFF has better performance than the L1-estimator, providing a better representation of learning a disturbance model.
We also observe the jitter in the prediction error of MPC~(L1-E) and MPC~(RFF). This is due to the stochasticity of control action determined by sampling-based MPC, which causes larger control action jitter. Since the aerodynamic effects depend on, \eg motor speeds, the control action jitter results in ground-truth disturbance jitter.

The plots of the ground-truth disturbances show the complexity of the disturbances used in our simulations. Specifically, in the case of a switching disturbance, the disturbance exhibits a sudden jump at $t=5s$; and in the case of a quadratic-phase disturbance, the disturbance changes faster as time increases. 
In addition, the $z-$axis disturbance exceeds the disturbance range of $[-2, \; 2] m/s^2$ used in DATT training. While for RFF, this disturbance can still be represented correctly within the training range of $\boldsymbol{\alpha} \sim \mathcal{N}(\boldsymbol{0}, \frac{1}{2}\boldsymbol{I})$ and $\sigma \in [0.001, \; 1]$, shown in \Cref{fig:sigma} and \Cref{fig:alpha}.

\newpage
\section{Results of Hardware Experiments}\label{sec:app-hw}
\begin{figure*}[h]
  \centering
  \includegraphics[width=\textwidth]{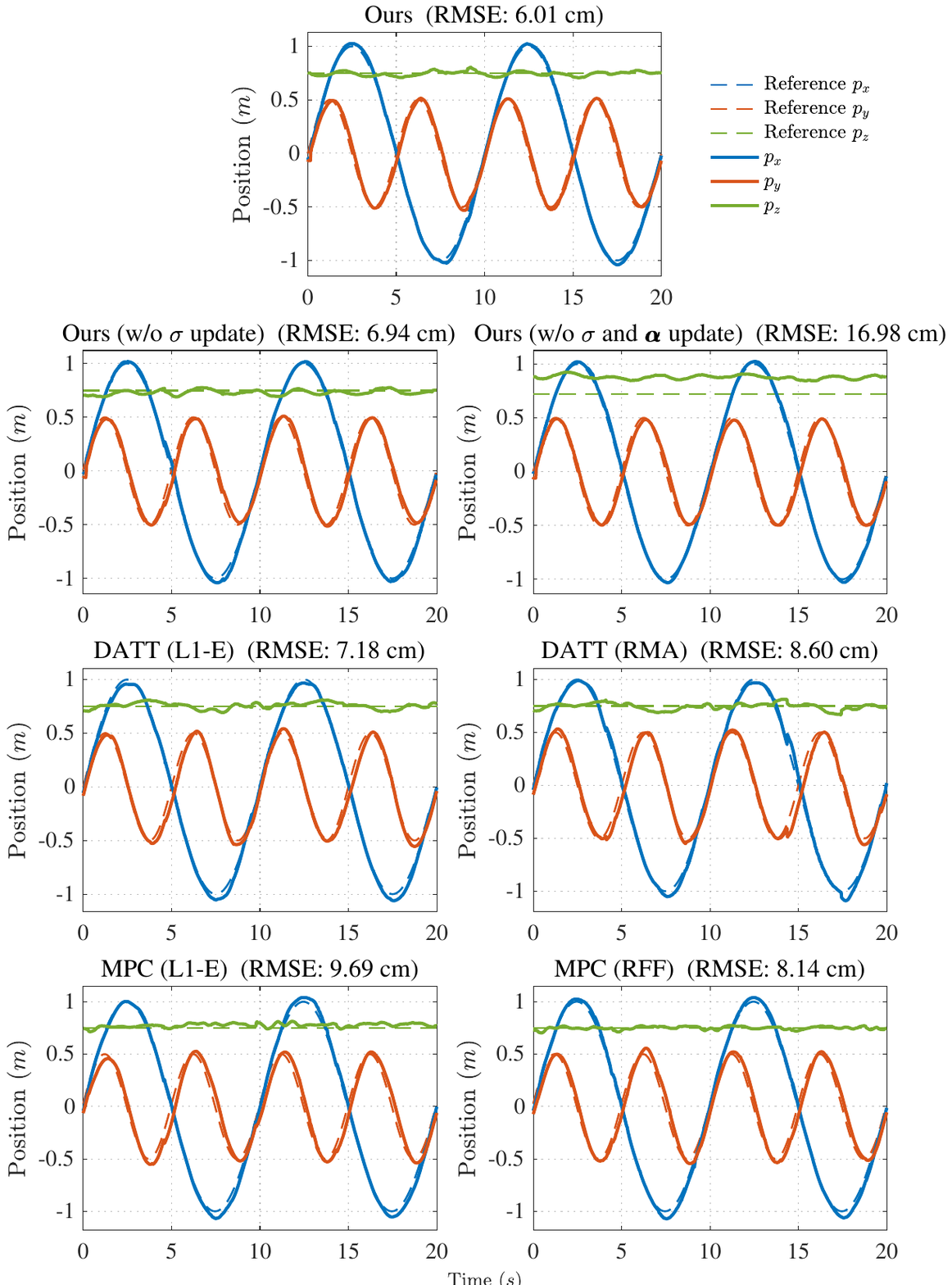}
  \caption{\textbf{Sample Trajectory of Hardware Experiment under  Ground Effect in \Cref{subsec:hw}.}
    Each subplot shows the $x-$, $y-$, $z-$ position and the reference trajectory (dashed) over a $20s$ evaluation window.
    Our method achieves the best tracking performance in RMSE.}
  \label{fig:traj-ground}
\end{figure*}

\begin{figure*}[t]
  \centering
  \includegraphics[width=\textwidth]{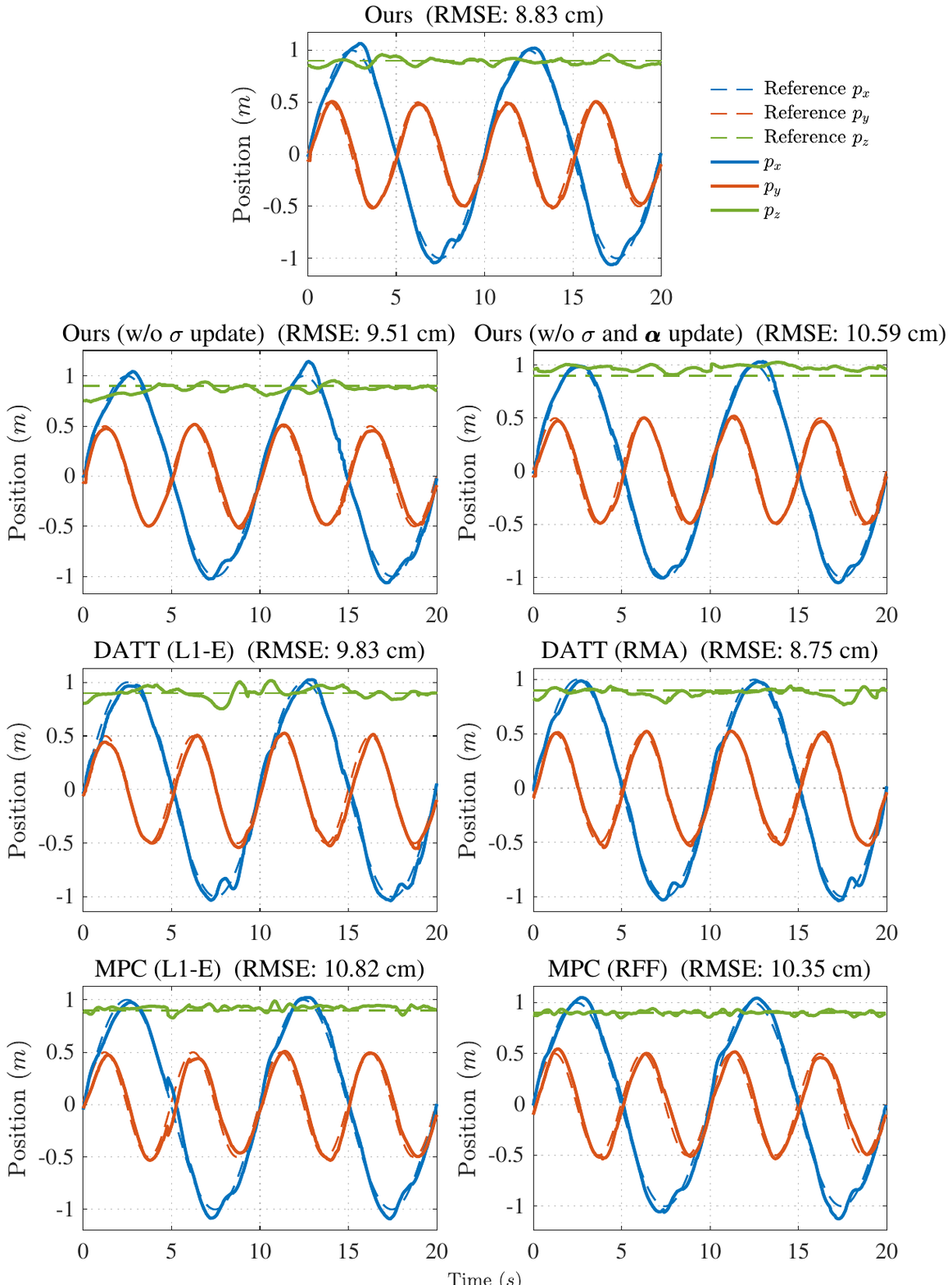}
  \caption{\textbf{Sample Trajectory of Hardware Experiment under Wind Disturbance in \Cref{subsec:hw}.}
    Each subplot shows the $x-$, $y-$, $z-$ position and the reference trajectory (dashed) over a $20s$ evaluation window.
    Our method achieves the best tracking performance in RMSE.}
  \label{fig:traj-wind}
\end{figure*}

\begin{figure*}[t]
  \centering
  \includegraphics[width=\textwidth]{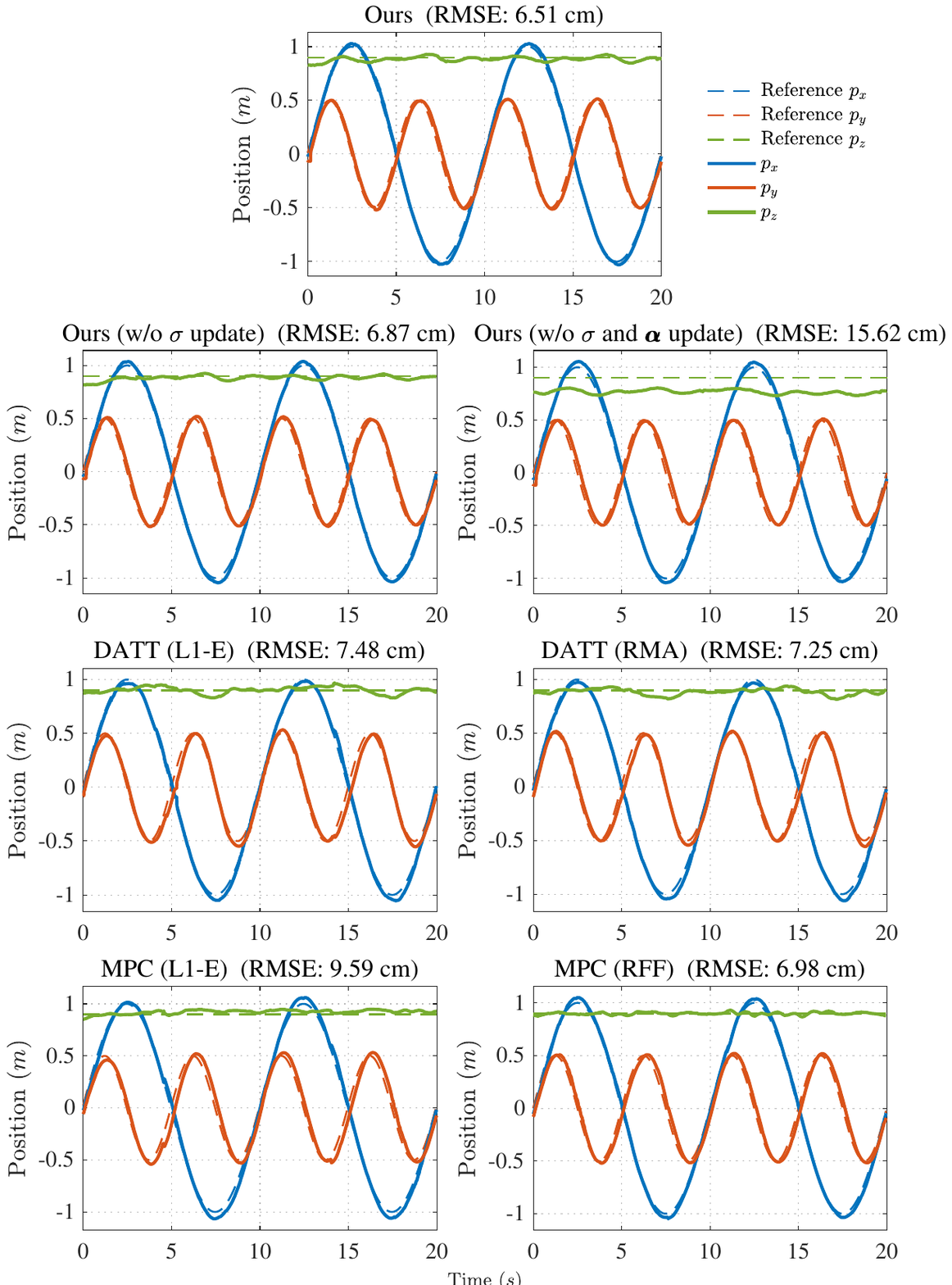}
  \caption{\textbf{Sample Trajectory of Hardware Experiment under Suspended Payload in \Cref{subsec:hw}.}
    Each subplot shows the $x-$, $y-$, $z-$ position and the reference trajectory (dashed) over a $20s$ evaluation window.
    Our method achieves the best tracking performance in RMSE.}
  \label{fig:traj-payload}
\end{figure*}

In this section, we provide sample trajectories of hardware experiments in \Cref{fig:traj-ground}, \Cref{fig:traj-payload}, and \Cref{fig:traj-wind}.

\end{appendices}

\end{document}